\documentclass[10pt,journal,compsoc]{IEEEtran}
\usepackage{fontenc}
\usepackage{graphicx}
\usepackage[cmex10]{amsmath}
\usepackage{array}
\usepackage{amssymb}
\usepackage{amsthm}
\usepackage{xcolor}
\usepackage[FIGBOTCAP]{subfigure}
\usepackage{mathrsfs}
\usepackage{bm}
\usepackage{bbm}
\usepackage{fullpage}
\usepackage[noadjust]{cite}
\usepackage{algorithm} 
\usepackage{algpseudocode}
\usepackage{multirow}
\usepackage{authblk}
\usepackage{caption}
\usepackage{color}

\newtheorem{lemma}{Lemma}
\newtheorem{definition}{Definition}

\newtheorem{theorem}{Theorem}

\newtheorem{proposition}{Proposition}

\begin{document}

\title{Robust Vertex Classification}
\author{Li Chen, Cencheng Shen, Joshua Vogelstein, Carey E. Priebe\thanks{L.C., C.S., and C.E.P are with Department of Applied Mathematics and Statistics, Johns Hopkins University. J.T.V. is with Department of Biomedical Engineering, Johns Hopkins University. }}

\IEEEtitleabstractindextext{%
\begin{abstract}
For random graphs distributed according to stochastic blockmodels, a special case of latent position graphs, adjacency spectral embedding followed by appropriate vertex classification is asymptotically Bayes optimal; but this approach requires knowledge of and critically depends on the model dimension. In this paper, we propose a sparse representation vertex classifier which does not require information about the model dimension. This classifier represents a test vertex as a sparse combination of the vertices in the training set and uses the recovered coefficients to classify the test vertex. We prove consistency of our proposed classifier for stochastic blockmodels, and demonstrate that the sparse representation classifier can predict vertex labels with higher accuracy than adjacency spectral embedding approaches via both simulation studies and real data experiments. Our results demonstrate the robustness and effectiveness of our proposed vertex classifier when the model dimension is unknown.

\end{abstract}

\begin{IEEEkeywords}
sparse representation, vertex classification, robustness, adjacency spectral embedding, stochastic blockmodel, latent position model, model dimension, classification consistency.
\end{IEEEkeywords}}

\maketitle

\section{Introduction}
Modern datasets have been collected with complex structures which contain interacting objects. Depending on the field of interest, such as sociology, biochemistry, or neuroscience, the objects can be people, organizations, genes, or neurons, and the interacting linkages can be communications, organizational positions, protein interactions, or synapses. Many useful models imply that objects sharing a ``class" attribute have similar connectivity structures. Graphs are one useful and appropriate tool to describe such datasets -- the objects are denoted by vertices and the linkages are denoted by edges. One interesting task on such datasets is vertex classification: determination of the class labels of the vertices. For instance, we may wish to classify whether a neuron is a motor neuron or a sensory neuron, or whether a person in a social network is liberal or conservative. 

In many applications, measured edge activity can be inaccurate, either missing or absolutely wrong, which leads to contaminated datasets. When the connectivity among a collection of vertices is invisible, occlusion contamination occurs. When we wrongly observe the connectivity among a collection of vertices, linkage reversion contamination occurs. The spectral embedding method on the adjacency matrix has been shown to be a valuable tool for performing inference on graphs realized from a stochastic blockmodel (\cite{sussman2012consistent}, \cite{fishkind2013consistent}, \cite{sussman2012}, \cite{tang2012universally}). One major issue is that such a method critically depends on a known model dimension, which is often unknown in practice. Moreover, for highly occluded graphs, classification composed with the spectral embedding method degrades in performance. 

This motivates us to propose a vertex classifier that does not require knowledge of the model dimension, yet achieves good performance for highly contaminated graphs. In this work, we apply the sparse representation classifier (\cite{wright2009robust}, \cite{wright2010sparse}, \cite{ShenChenPriebe2015}) to do vertex classification on graph data, which performs well in object recognition with contamination and does not require dimension selection. In particular, we provide both theoretical performance guarantee of this classifier for the stochastic blockmodel, and its numerical advantages via simulations and various real graph datasets. Furthermore, the proposed classifier maintains low misclassification error under both occlusion and linkage reversion contamination.

This paper is organized as follows: in Section \ref{sec:bg_fr}, we provide background on the classification framework, review the latent position model and the stochastic blockmodel, and present the vertex classification framework. In Section \ref{sec:motivation}, we describe the motivation for investigating robust vertex classification, and propose two contamination models on stochastic blockmodels. In Section \ref{sec:src}, we propose a sparse representation classifier for vertex classification and prove consistency of our proposed classifier for the stochastic blockmodel under certain condition on the model parameters. In Section \ref{sec:experiments}, we demonstrate the effectiveness of our proposed classifier via both simulated and real data experiments. In Section \ref{discussion}, we discuss the practical advantages of applying sparse representation classifier to graphs. All theoretical proofs 
are in the supplementary material. 

\section{Background and Framework} \label{sec:bg_fr}
\subsection{Classification in the Classical Setting}
Let $[K] = \{1, ..., K\}$ for any positive integer $K$. Let $(X, Y) \sim F_{XY}$, where the feature vector $X$ is an $\mathbb{R}^{d}$-valued random vector, $Y$ is a $[K]$-valued class label, and $F_{XY}$ is the joint distribution of $X$ and $Y$. Let $\pi_{k} = P(Y = k)$ be the class priors and let $g: \mathbb{R}^{d} \rightarrow [K]$ provide one's guess of $Y$ given $X$, for which $g$ is a classifier. We intend to classify a test observation $X$ -- that is, estimate its true but unknown label $Y$ via $g(X)$. An error occurs when $g(X) \neq Y$, and the probability of error is denoted by $L(g) = P(g(X) \neq Y)$. The optimal classifier is defined by $g^{*} = \arg\min_{g:\mathbb{R}^{d} \rightarrow [K]}P(g(X) \neq Y)$, which is the \textit{Bayes} classifier achieving the minimum possible error. In the classical setting of supervised learning, we observe training data $\mathcal{T}_{n} = \{(X_{1},Y_{1}),...,(X_{n},Y_{n})\} \overset{iid}{\sim} F_{XY} $. The performance of $g_{n}$ is measured by the conditional probability of error defined by
\begin{equation*}
L_{n} = L(g_{n}) = P(g_{n}(X; \mathcal{T}_{n}) \neq Y |\mathcal{T}_{n}),
\end{equation*}
for a sequence of classifiers $\{g_n, n\geq 1\}$. The sequence of classifiers is consistent if $\lim_{n \rightarrow \infty} L_{n} \rightarrow L^{*}$ as $n \rightarrow \infty$; and it is universally consistent if $\lim_{n \rightarrow \infty} L_{n} = L^{*}$ with probability $1$ for any distribution $F_{XY}$ \cite{devroye1996probabilistic}.

\subsection{Vertex Classification in the Random Graph Setting}
This supervised learning framework is adapted for the setting of random graphs. A graph is a pair $G = (V , E)$ consisting of a set of vertices or nodes $V = [n] = \{1, 2, ..., n\}$ and a set of edges $E \subset$  $  [n] \choose 2$. In this work, we assume that all graphs are simple; that is, the graphs are undirected, unweighted, and non-loopy. The adjacency matrix of $G$, denoted by $A$, is $n$-by-$n$ symmetric, binary, and hollow, i.e., the diagonals of $A$ are all zeros. Each entry $A_{uv} =  A_{vu} = 1$, if there is an edge between vertices $u$ and $v$; $A_{uv} = 0$ otherwise \cite{west2001introduction}. A random graph is a graph-valued random variable $\mathbb{G}: \Omega \rightarrow \mathcal{G}_{n}$, where $\Omega$ denotes the probability space, and $\mathcal{G}_{n}$ the collection of all possible $2^{n \choose 2}$ graphs on $V = [n]$. For instance, one frequently occurring random graph model is the so-called Erdos-Renyi graph, $\mathcal{ER}(n,p)$, in which each pair of vertices has an edge independently with probability $p$ \cite{bollobas2001random}.

Our exploitation task is vertex classification. We observe the adjacency matrix $A \in \{0,1\}^{(n+1) \times (n+1)}$ on $n+1$ vertices $\{v_1,  \ldots, v_n, v\}$ and the class labels $Y_i \in [K]$ associated with the first $n$ vertices. Our goal is to estimate the class label $Y$ of the test vertex $v$ via a classifier $g: \{0,1\}^{(n+1) \times (n+1)} \rightarrow [K]$ such that the probability of error $P(g(A) \neq Y)$ is small.

\subsection{Related Work and Definitions}
In our setting, we describe the stochastic blockmodel and vertex classification from the perspective of a latent position graph framework. Hoff et al.\ \cite{hoff2002latent} proposed a latent position graph model. In this model, each vertex $v$ is associated with an unobserved latent random vector $X_{v}$ drawn independently from a specified distribution $F$ on $\mathbb{R}^{d}$. The adjacency matrix entries $A_{uv}|(X_{u}, X_{v}) \sim \text{Bernoulli}(l(X_{u}, X_{v}))$ are conditionally independent, where $l: \mathbb{R}^{d} \times \mathbb{R}^{d} \rightarrow [0,1]$ is the link function. The random dot product graph model proposed in \cite{young2007random} is a special case of the latent position model, where the link function $l(X_{u}, X_{v})$ is the inner product of latent positions, $l(X_{u}, X_{v}) = \langle X_{u}, X_{v}\rangle$. For the purpose of theoretical analysis and simulation in this paper, we mainly consider the stochastic blockmodel introduced in \cite{holland1983stochastic}, which is a random graph model with a set of $n$ vertices randomly drawn from $K$ block memberships. Conditioned on the $K$-partition, edges between all the pairs of vertices are independent Bernoulli trials with parameters determined by the block memberships. 

Below we formally present the definitions of the latent position model and the stochastic blockmodel, which provide the framework for our exploitation task of vertex classification.
\begin{definition}
\textbf{Latent Position Model (LPM) } \normalfont Let $F$ be a distribution on $[0, 1]$, $X_{1},..., X_{n} \overset{iid}\sim F$, and define $ Z := [X_{1},..., X_{n}]^{T} \in \mathbb{R}^{n \times d}$. Suppose rank$(Z)=d$, and denote $\mathbf{P} \in [0, 1]^{n \times n}$ as the communication probability matrix, where each entry $\mathbf{P}_{ij} $ is the probability that there is an edge between vertices $i$, $j$ conditioned on $X_{i}$ and $X_{j}$. Let $A \in \{0, 1\}^{n \times n}$ be the random adjacency matrix. Then $(Z, A) \sim LPM(F)$ if and only if the following conditional independence relationship holds:
\begin{equation}
P(A | X_{1}, ..., X_{n}) = \Pi_{i < j}\mathbf{P}_{ij}^{A_{ij}}(1-\mathbf{P}_{ij})^{1-A_{ij}},
\end{equation}
\begin{equation}
\mathbf{P}_{ij} = P(A_{ij} =1| X_{i}, X_{j}).
\end{equation} 
The $X_{i}'s$ are the latent positions for the model, and the rank of the communication probability matrix $\mathbf{P}$ satisfies $\text{rank}(\mathbf{P}) \leq d$. 
\end{definition}

\begin{definition} \textbf{Stochastic Blockmodel (SBM)} 
\normalfont Let $K$ be the number of blocks, and $\pi$ be a length $K$ vector in the unit simplex $\Delta^{K-1}$. The block memberships of the vertices are given by $Y(v) \overset{iid}\sim \text{Multinomial}([K], \pi)$. Let $B$ be a $K \times K$ symmetric matrix specifying block communication probabilities. Then $A \sim SBM([n], B, \pi)$ if and only if the following conditional independence relationship holds:
\begin{align*}
\mathbf{P}_{ij} & = P(A_{ij} =1| X_{i}, X_{j}) = P(A_{ij} =1| Y_{i}, Y_{j}) \\
&= B_{Y_{i}, Y_{j}}. 
\end{align*}
Note that SBM is a special case of LPM, because the latent positions of an SBM are mixtures of the point masses, which are the eigenvectors of $B$. The unknown latent positions $X_i$ and $X_j$ of vertices $i$, $j$ determine their memberships $Y_i$ and $Y_j$. And for vertex classification on SBM, the Bayes error $L^*= 0$ \cite{sussman2012consistent}.

\end{definition}
\begin{definition}\textbf{Model Dimension}
\normalfont For stochastic blockmodels, the model dimension refers to the rank of the communication probability matrix. A $d$-dimensional SBM satisfies $\text{rank}(\mathbf{P})=  \text{rank}(B) = d $, for which $d \leq K$; if $B$ is full rank, then $d=K$.
\end{definition}
\begin{definition}\textbf{Adjacency Spectral Embedding in Dimension $\hat{d}$} 
\normalfont
Let $A$ be defined as in Definition 1. Let $A = U_{A}S_{A}U_{A}^{T}$ be the full spectral decomposition of $A$, where $S_A = \text{Diag}(\lambda_1, \lambda_2, \ldots, \lambda_n)$ with $\lambda_1 \geq \lambda_2 \geq \dots \geq \lambda_n$. Let $S_{A, \hat{d} } = \text{Diag}(\lambda_1, \lambda_2, \ldots, \lambda_{\hat{d}}) \in \mathbb{R}^{\hat{d} \times \hat{d}}$, containing the $\hat{d}$ largest eigenvalues of $A$. Let $U_{A, \hat{d}} \in \mathbb{R}^{n \times \hat{d}}$ be the matrix containing the corresponding eigenvectors as its column vectors. The estimate of latent positions of SBM via adjacency spectral embedding in dimension $\hat{d}$ is defined as $\hat{Z}_{\hat{d}} = U_{A, \hat{d}}S_{A, \hat{d} }^{\frac{1}{2}}$, for $1\leq \hat{d} \leq n$. We denote the method of adjacency spectral embedding to dimension $\hat{d}$ as ASE$_{\hat{d}}$.
\end{definition}

Many techniques have been developed to infer the latent positions via the realized adjacency matrix. Bickel et al.\ \cite{bickel2011method} used subgraph counts and degree distributions to consistently estimate stochastic blockmodels. Sussman et al.\ \cite{sussman2012consistent} proved the consistency of spectral partitioning on the adjacency matrix of stochastic blockmodels. Rohe et al.\ \cite{rohe2011spectral} proved a consistent spectral partitioning procedure on the Laplacian of the stochastic blockmodels. Fishkind et al.\ \cite{fishkind2013consistent} showed the consistency of adjacency spectral partitioning, when the model parameters are unknown. Athreya et al.\ \cite{athreya2013limit} proved a central limit theorem for the adjacency spectral embedding of stochastic blockmodels. 

In the area of clustering and classification, there exists intensive works regarding unsupervised learning for graph data \cite{handcock2007model}, \cite{lei2014consistency}, \cite{chaudhuri2012spectral}, \cite{chen2012clustering}, \cite{balakrishnan2011noise}, \cite{lyzinski2014perfect} and \cite{chen2014stochastic}; as well as supervised learning, such as \cite{sussman2012}, \cite{tang2013universally}, \cite{priebe2014statistical} for vertex classification, and \cite{fishkind2013vertex} and \cite{ChenPhd2015} for vertex nomination. 

Our task in this paper is vertex classification. However, we do not and can not observe the latent positions $X_{1},...,X_{n}, X$; otherwise, we are back in the classical setting of supervised learning. We assume that the class-conditional density $X_i|Y_i =k \sim f_k$ with class priors $\pi$ as before, that is, $P(Y_i = k|X_i = x) = \frac{\pi_k f_k(x)}{\sum_{j \in [K]} \pi_j f_j(x)}$. We denote the test vertex as $v$ whose latent position is $X$, and we shall assume that we do not observe the label $Y$. 

\section{Motivation}\label{sec:motivation} 
\begin{figure}[!ht]  
  \centering
    \includegraphics[width=2.5in]{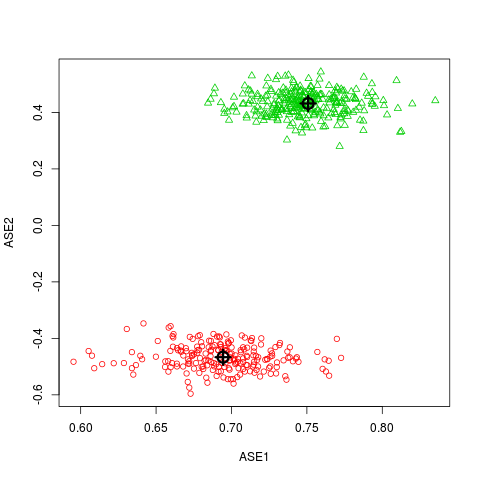}          
    \caption{\textbf{An example of adjacency spectral embedding.} Example of adjacency spectral embedding (ASE$_{d = 2}$) with $n=500$. The parameters $B$ and $\pi$ are given in Equation \ref{eq:simParams}. The latent position of this SBM is a mixture of point masses at $X_1 = (0.695, -0.467)^T$ and $X_2 = (0.751, 0.432)^T$.} 
    \label{fig:motivating_example}
\end{figure}

Our motivation for proposing a robust vertex classifier comes from asking the question: how well can vertex classifiers perform when model assumptions do not hold. If the model dimension $d$ is known or can be estimated correctly, ASE$_{d}$ consistently estimates the latent positions for SBM \cite{sussman2012consistent}. Figure \ref{fig:motivating_example} presents an example of ASE$_{d}$, where vertices from two classes are well separated in the embedded space. A subsequent $k$-nearest neighbor ($k$NN) classifier on ASE$_{d}$ is universally consistent for SBM \cite{sussman2012}. That means regardless of what distribution the latent positions are drawn from, $k$NN$\circ$ASE$_{d}$ achieves the Bayes error $L^{*}$ asymptotically as $k \rightarrow \infty$, $n \rightarrow \infty$ and $k/n \rightarrow 0$. In particular, for stochastic blockmodels, $1$NN$\circ$ASE$_{d}$ is asymptotically Bayes optimal \cite{sussman2012consistent}. 

Athreya et al.\ \cite{athreya2013limit} proved a central limit theorem that for $K$-block and $d$-dimensional SBM, $\hat{Z}_{d}$ via ASE$_{d}$ is distributed asymptotically as a $K$-mixture of $d$-variate normal with covariance matrices of order $\frac{1}{n}$. This asymptotic result holds true for any constant $K$, any finite $d$, all but finitely many $n$, and does not require equal number of vertices per partition. This result implies that quadratic discriminant analysis (QDA) and linear discriminant analysis (LDA) on the represented data $\hat{Z}_{d}$ of stochastic blockmodels are asymptotic Bayes plug-in classifiers,
while LDA requires a fewer number of parameters to fit. Hence in our analysis, we employ two consistent classifiers 1NN$\circ$ASE$_{d}$ and LDA$\circ$ASE$_{d}$ for vertex classification on stochastic blockmodels. 

Importantly, having information on the model dimension $d$ is critical to adjacency spectral approaches. When $d$ is given, ASE$_{d}$ is consistent, and 1NN$\circ$ASE$_{d}$, LDA$\circ$ASE$_{d}$ are asymptotically Bayes optimal. When $d$ is not known, Sussman et al.\ \cite{sussman2012consistent} estimates $d$ via a consistent estimator. However, for the consistent estimator to be accurate, the required number of vertices $n$ will depend highly on the graph density, and increases rapidly as the expected graph density decreases. Fishkind et al.\ \cite{fishkind2013consistent} shows that if we pick a positive integer $\hat{d} \geq d$, then ASE$_{\hat{d}}$ is still consistent as $n \rightarrow \infty$. However, for a finite number of vertices, 1NN$\circ$ASE$_{\hat{d}}$ and LDA$\circ$ASE$_{\hat{d}}$ degrade significantly in performance compared to 1NN$\circ$ASE$_{d}$ and LDA$\circ$ASE$_{d}$. Moreover their performance on real data can be very sensitive to the choice of embedding dimension. Our focus is on removing the need to know the model dimension $d$ and still maintaining low error rate for vertex classification, so the classification procedure can be robust and suitable for practical inference when the model assumptions do not hold. 

\subsection{Contamination Procedures}
To assess the robustness of the vertex classifiers for stochastic blockmodels, we propose two scenarios of contamination that change the model dimension of SBM. Suppose the uncontaminated graph model $\mathbb{G}_{\text{un}}$ is a stochastic blockmodel $\mathbb{G}_{\text{un}} \sim \text{SBM}([n], B_{\text{un}}, \pi_{\text{un}})$. Denote the communication probability matrix of $\mathbb{G}_{\text{un}}$ as $\mathbf{P}_{\text{un}}$. We can write $\mathbf{P}_{\text{un}} = Z_{\text{un}}Z_{\text{un}}^T$, where $Z_{\text{un}}$ is the latent positions of the uncontaminated model \cite{sussman2012consistent}, and suppose rank$(B_{\text{un}}) = d$. Denote by $\delta_{i}(M)$ the $i$-th largest singular value of a matrix $M$. 

\subsubsection{Contamination I: The Occlusion Model}
Let $p_{o} \in [0,1]$ denote the occlusion rate. We randomly select $(100p_o)\%$ vertices out of the $n$ vertices and set the probability of connectivity among the selected vertices to be $0$. In this scenario, the probability of connectivity between the contaminated vertices and the uncontaminated vertices remains the same as in $\mathbb{G}_{\text{un}}$. This occlusion procedure can be formulated as a stochastic blockmodel $\mathbb{G}_{\text{occ}}$ with the following parameters:
\begin{equation}
B_{\text{occ}} =
\left( {\begin{array}{cc}
 B_{\text{un}} & B_{\text{un}} \\
 B_{\text{un}} & 0_{K \times K}
 \end{array} } \right) \in \mathbb{R}^{2K \times 2K}, 
\end{equation}
\begin{equation}
\pi_{\text{occ}} =  [(1-p_{o})\pi_{\text{un}}^{T}, \enspace p_o\pi_{\text{un}}^{T}]^{T} \in \mathbb{R}^{2K }.
\end{equation}

Denote the communication probability matrix of $\mathbb{G}_{\text{occ}}$ by $\mathbf{P}_{\text{occ}}$. It always holds that $\delta_{1}(\mathbf{P}_{\text{occ}}) \leq \delta_{1}(\mathbf{P}_{\text{un}}) \leq n$, and it almost always holds that rank$(B_{\text{occ}}) =$ rank$(\mathbf{P}_{\text{occ}})=2d$. That is, the true model dimension of the occluded graph is $2d$ instead of $d$. The proofs to the above claims are provided in the supplementary material. 

Both $B_{\text{occ}}$ and $\mathbf{P}_{\text{occ}}$ have $d$ positive and $d$ negative eigenvalues, where the $d$ negative eigenvalues are due to occlusion contamination. The number of blocks in the contaminated model $\mathbb{G}_{\text{occ}}$ rises to $2K$, where $K$ blocks correspond to $(1-p_{o})\pi_{\text{un}}$ and the other $K$ blocks correspond to $p_{o}\pi_{\text{un}}$. Although the number of blocks in the model changes to $2K$ due to contamination, the number of classes in the vertex classification problem remains $K$. As $p_{o} \rightarrow 1$, the number of contaminated vertices approaches $n$, indicating that the majority of the edges are sampled from the contamination source $0_{K \times K}$; as a result, the adjacency matrix $A$ becomes sparser and sparser.

Note that our occlusion scenario randomly selects the vertices; and conditioned on selecting the contaminated vertices, the edges between these vertices are missing deterministically. Therefore the edges are not missing completely at random in this occlusion contamination procedure. 

\subsubsection{Contamination II: The Linkage Reversion Model}
Let $p_{l} \in [0, 1]$ denote the linkage reversion rate. We randomly select $(100p_{l})\%$ vertices out of the $n$ vertices and reverse the connectivity among all the selected vertices. The probability of connectivity between the contaminated vertices and the uncontaminated vertices remains the same as in $\mathbb{G}_{\text{un}}$. The linkage reversion contamination can be formulated as a stochastic blockmodel $\mathbb{G}_{\text{rev}}$ with the following parameters:
\begin{equation}
B_{\text{rev}} =
\left( {\begin{array}{cc}
 B_{\text{un}} & B_{\text{un}} \\
 B_{\text{un}} & J_{K \times K} - B_{\text{un}}
 \end{array} } \right) \in \mathbb{R}^{2K \times 2K}, 
\end{equation}
\begin{equation}
\pi_{\text{rev}} =  [(1-p_{l})\pi_{\text{un}}^{T},  \enspace p_{l}\pi_{\text{un}}^{T}]^{T} \in \mathbb{R}^{2K }.
\end{equation}
The matrix $J_{K \times K} \in \mathbb{R}^{K \times K}$ is the matrix of all ones. Denote the communication probability matrix of $\mathbb{G}_{\text{rev}}$ by $\mathbf{P}_{\text{rev}}$. If $\text{rank}(B_{\text{un}}) = d$, then it almost always holds that $d+1 \leq \text{rank}(B_{\text{rev}}) =  \text{rank}(\mathbf{P}_{\text{rev}})\leq 2d$, since the block matrix $J_{K \times K} - B_{\text{un}}$ has rank at most $d$. The number of blocks in the contaminated model also increases to $2K$, similar to the occlusion model. As $p_{l} \rightarrow 1$, we recover the complement of $\text{SBM}([n], B_{\text{un}}, \pi_{\text{un}})$ -- that is, $\text{SBM}([n], J_{K \times K} - B_{\text{un}}, \pi_{\text{un}})$.

\subsection{The Contamination Effect}
When the stochastic blockmodels are contaminated by the above two procedures, the model parameters and the model dimension are changed. Suppose both the original model dimension $d$ and the contamination information are known, then we can use the contaminated model dimension $d_{\text{occ}}=2d$ or $d_{\text{rev}} \in [d+1, 2d]$ for embedding, so that ASE$_{d_{\text{occ}}}$ and ASE$_{d_{\text{rev}}}$ followed by 1NN and LDA are asymptotically Bayes optimal. However, if we only know the contamination but not the model dimension, then adjacency spectral embedding will require the estimation of an embedding dimension; and if we know $d$ but not the contamination, we usually consider $d$ as the default embedding dimension. In either case, the embedding dimension used may not be the best choice for adjacency spectral embedding and subsequent classification.  

\begin{figure}[!ht]  
  \centering
    \includegraphics[width=3.in]{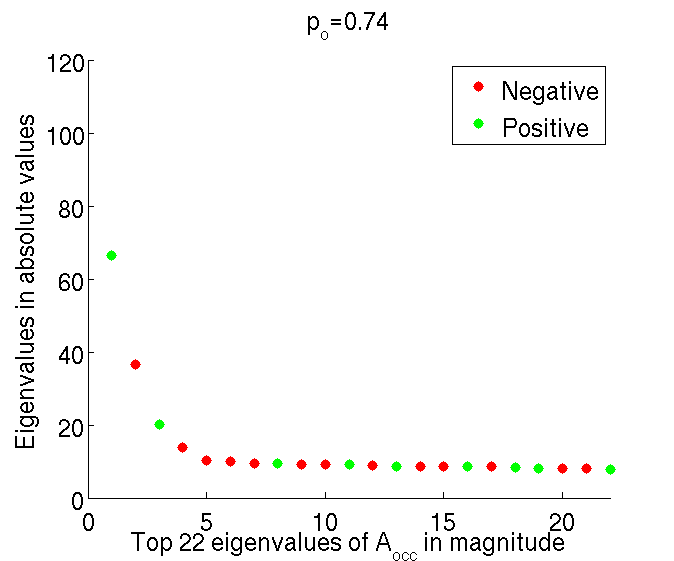}   
    \caption{\textbf{Scree plot of the occlusion contaminated adjacency matrix.} Scree plot of the occlusion contaminated adjacency matrix $A_{\text{occ}}$ at occlusion rate $p_o =0.74$ with $n=200$. The parameters $B_{\text{un}}$ and $\pi_{\text{un}}$ are given in Eq.\ \ref{eq:simParams}. The red dots are the negative eigenvalues of $A_{\text{occ}}$ due to occlusion contamination, and the green dots are the positive eigenvalues of $A_{\text{occ}}$. Profile likelihood \cite{zhu2006automatic} method always suggests $\hat{d}=2$ for this scree plot. }
    \label{fig:ase_scree1}
\end{figure}

\begin{figure}[!ht]  
  \centering
     \includegraphics[width=3in]{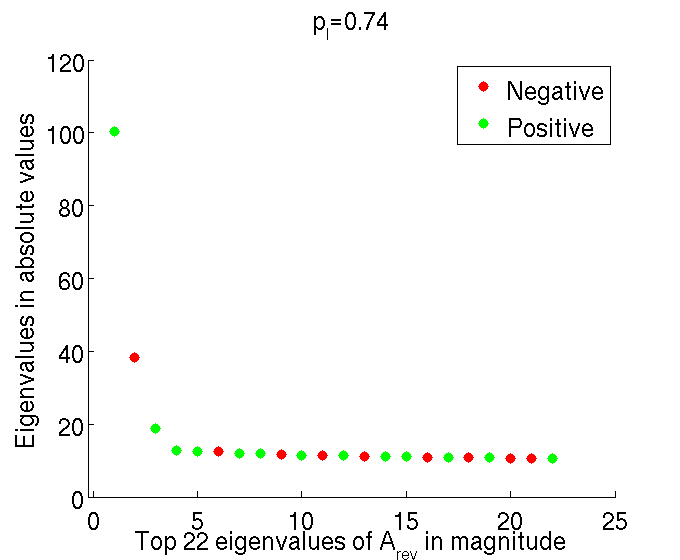} \\
    \caption{\textbf{Scree plot of the linkage reversion contaminated adjacency matrix.} Scree plot of the linkage reversion contaminated adjacency matrix $A_{\text{rev}}$ at linkage reversion rate $p_l =0.74$ with $n=200$. The parameters $B_{\text{un}}$ and $\pi_{\text{un}}$ are given in Eq.\ \ref{eq:simParams}. The red dots are the negative eigenvalues of $A_{\text{occ}}$ due to linkage reversion, and the green dots are the positive eigenvalues. Profile likelihood method \cite{zhu2006automatic} always suggests $\hat{d}=2$ for this scree plot. }
    \label{fig:ase_scree2}
\end{figure}

Figure \ref{fig:ase_scree1} and Figure \ref{fig:ase_scree2} provide two examples of the scree plots obtained from the contaminated adjacency matrices $A_{\text{occ}}$ and $A_{\text{rev}}$, for which the original model dimension is $d=2$. Using $d=2$ is clearly not the best choice in the contaminated data; and if we decide to estimate $d$, this remains a very challenging task, despite various procedures and criteria for dimension selection \cite{jackson2005user}. Here we use a principled automatic dimension selection procedure using the profile likelihood by \cite{zhu2006automatic}, to estimate the embedding dimension based on the scree plot. 

However, in the setting of Figure \ref{fig:ase_scree1} and Figure \ref{fig:ase_scree2}, Monte Carlo investigation yields $\hat d=2$ every time as the elbow (500 times out of 500 Monte Carlo replicates), using the full spectrum or a partial spectrum of the largest 22 eigenvalues in magnitude respectively. The second elbow selected by \cite{zhu2006automatic} concentrates around 80 and 11 using the full spectrum and the partial spectrum respectively. Even though $\hat d = 3, 4$ are better for classification purpose in these two contaminated graphs, they are not selected by the dimension selection method of \cite{zhu2006automatic}.

\begin{figure}[!ht]  
  \centering
    \includegraphics[width=3.in]{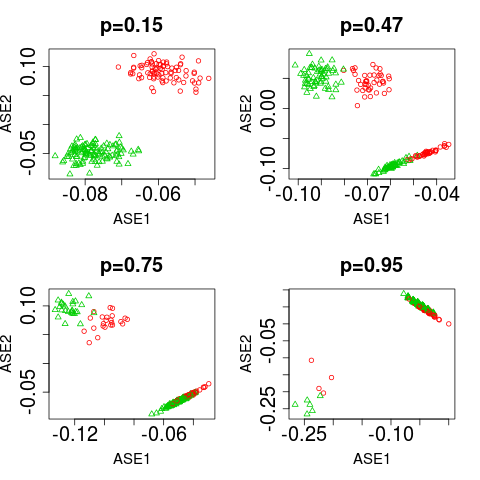}     
    \caption{\textbf{The occlusion contamination effect on estimated latent positions.} A depiction of the occlusion effect on the latent positions as reflected in the estimated latent positions $\hat{Z}_{\hat{d}=2}$ with $n=200$. The parameters $B_{\text{un}}$ and $\pi_{\text{un}}$ are given in Eq. \ref{eq:simParams}. The four-panel displays the latent position estimation for different occlusion rate $p_o$. As $p_{o}$ increases, vertices from different blocks become close in the embedded space. For $p_{o}$ close to $1$, ASE$_{\hat{d}=2}$ will eventually yield only one cloud at $0$.  }   
    \label{fig:ase_occlusion}
\end{figure}

\begin{figure}[!ht]  
  \centering
    \includegraphics[width=3.in]{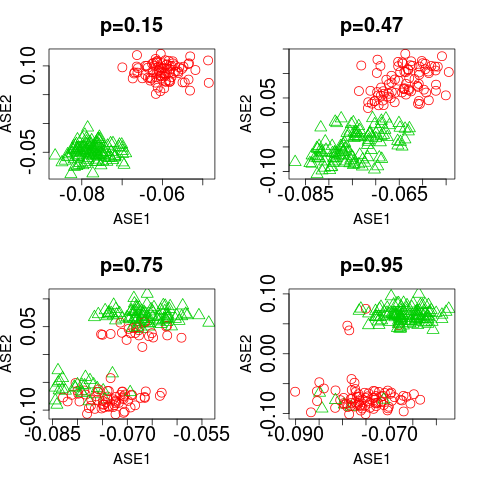}   
    \caption{\textbf{The linkage reversion contamination effect on estimated latent positions.} A depiction of the linkage reversion effect on the latent positions as reflected in the estimated latent positions $\hat{Z}_{\hat{d}=2}$ with $n=200$. The parameters $B_{\text{un}}$ and $\pi_{\text{un}}$ are given in Eq. \ref{eq:simParams}. The four-panel displays the latent position estimation for different linkage reversion rate $p_l$. As $p_{l}$ increases, vertices from different blocks become close in the embedded space. For $p_{l}=1$, ASE$_{\hat{d}=2}$ will yield two clouds corresponding to SBM($200$, $J_{2\times2}-B_{\text{un}}$, $\pi_{\text{un}}$). } 
    \label{fig:ase_flip}
\end{figure}

Notwithstanding the results in \cite{sussman2012} and \cite{fishkind2013consistent}, we cannot be guaranteed to successfully choose the embedding dimension in practice. Consequently, the performance of ASE method and subsequent classification will suffer. Figure \ref{fig:ase_occlusion} and Figure \ref{fig:ase_flip} demonstrate that, as the contamination proportions $p_o$ and $p_l$ increase, latent positions change as reflected in the estimated latent positions $\hat{Z}_{\hat{d}=2}$  and $\hat{Z}_{\hat{d}=2}$ plots, for which the profile likelihood method always yield $\hat{d}=2$ for the contaminated data. In particular, as the occlusion rate $p_o$ increases, more vertices from different classes are embedded close together.

Furthermore, vertex classification on the contaminated $\hat{Z}_{\hat{d}}$ using 1NN or LDA will degrade in performance, as illustrated later in the simulation and Figure \ref{fig:err_occlusion}. Indeed, the model dimension critically determines the success of vertex classification based on the ASE procedures, whereas in practice, the model dimension is usually unknown. This motivates us to seek a robust vertex classifier which does not heavily depend on the model selection and still attains good performance.

\section{The Sparse Representation Classifier for Vertex Classification} \label{sec:src}
In this section, we propose to use the sparse representation classifier (SRC) for robust vertex classification. Instead of employing adjacency spectral embedding and applying subsequent classifiers on $\hat{Z}_{\hat{d}}$, we recover a sparse representation of the test vertex with respect to the vertices in the training set, and use the recovered sparse representation coefficients to classify the test vertex. 
	
For the purpose of algorithm presentation, in this section we slightly abuse the notation to denote $A$ as the adjacency matrix on the training vertices $\{v_1, \ldots, v_n\}$ with known labels $Y_i \in [K]$, and denote $\phi$ as the adjacency column with respect to the testing vertex $v$ with an unknown label $Y$; note that this is almost equivalent to let $A$ be the adjacency matrix for $\{v_1, \ldots, v_n, v\}$ as in previous sections, then split the first $n$ columns for training and the last column for testing, except the last row is not used. 

Now suppose there are $n_k$ training vertices in each class $k$, so that $n = \sum_{k \in [K]}n_k$. Let $a_{k,1},...,a_{k,n_{k}}$ denote the columns in $A$ corresponding to the $n_k$ training vertices in class $k$. Define a matrix $D_k=[d_{k,1},...,d_{k,n_{k}}] \in \mathbb{R}^{n \times n_{k}}$, where each column $d_{k,j} = \frac{a_{k,j}}{\|a_{k,j}\|_2}$ for $1 \leq j \leq n_k$; then we concatenate $D_1, \ldots, D_K$ such that $D := [D_1, \ldots, D_K] \in \mathbb{R}^{n \times n}$. Namely the matrix $D$ re-arranges the columns of $A$ by classes, and normalize each column to have $\ell 2$ unit norm. 

Also normalize $\phi$ to unit norm. Then SRC is applied to $D$ and $\phi$ directly, by first solving the $\ell 1$-minimization problem
\begin{equation} 
\arg\min \| \beta \|_{1} \text{ subject to } \phi=D \beta+\epsilon,
\end{equation}
followed by subsequent classification on the sparse representation $\beta$. This procedure does not require spectral embedding of the adjacency matrix, and is originally used by \cite{wright2009robust} to do robust face recognition.

In subsection~\ref{SRCSection1} we show the algorithmic and implementation details, and argue why SRC is applicable for graphs; then a consistency result of SRC for the stochastic blockmodel is proved, followed by relevant discussions in subsection~\ref{SRCSection2}.

\subsection{The Algorithm}
\label{SRCSection1}
The algorithm is summarized in Algorithm 1. The only computational costly step in Algorithm 1 is $\ell 1$ minimization. Many algorithms, such as $\ell 1$ homotopy \cite{DonohoTsaig2008}, augmented Lagrangian multiplier \cite{yang2011orthonormal}, orthogonal matching pursuit \cite{tropp2004greed}, etc., are developed to solve $\ell 1$ minimization. In this paper, we use orthogonal matching pursuit (OMP) to solve Equation~\ref{l1min}, which is a fast approximation of exact $\ell 1$ minimization; details of various $\ell 1$ minimization and OMP are available in \cite{DonohoTsaig2008}, \cite{tropp2007signal}, \cite{yang2011orthonormal}, and \cite{ShenChenPriebe2015}. 

\begin{algorithm}
\caption{Robust vertex classification.} 
\begin{algorithmic} 
\State \textbf{Goal}: Classify the vertex $v$ whose unknown label is $Y$. 
\State \textbf{Input}: Adjacency matrix $A \in \{0,1\}^{(n) \times (n)}$ from the training vertices $\{v_1, \ldots, v_n\}$, where each column $a_{i}$ contains the adjacency column of $i$th vertex to all other training vertices, and all vertices are associated with observed labels $Y_i \in [K]$. Let $\phi \in \{0,1\}^n$ be the testing vertex containing its connectivity to all training data. 
\State 1. \textbf{Arrange and scale all vertices}: Re-arrange columns of $A$ in class order, and normalize the column to $\ell 2$ unit norm. Denote the resulting matrix as $D$. Also scale the testing adjacency column $\phi$ to have unit norm.

\State 2. \textbf{Find a sparse representation of $\phi$ by $\ell 1$ minimization:}
\begin{equation}
\label{l1min}
\hat{\beta} = \arg\min \|\beta\|_{1} \text{ subject to } \phi = D \beta +\epsilon.  \nonumber
\end{equation}

\State 3. \textbf{Compute the distance of $\phi$ to each class $\textit{k}$:} $r_{k}(\phi) = \|\phi - D\hat{\beta}_{k}\|_{2}$, where $\hat{\beta}_{k} = [0, ..., 0, \hat{\beta}_{k,1}, ..., \hat{\beta}_{k,n_{k}}, ..., 0] \in \mathbb{R}^{n}$ is the recovered coefficients corresponding to the $k$-th class. 

\State 4. \textbf{Classify test vertex}: $\hat{Y} = \arg\min_{k}r_{k}(\phi)$.				    						 				  
\end{algorithmic} 
\label{alg:src}
\end{algorithm}  

Usually there is a model selection parameter for stopping $\ell 1$ minimization, namely the noise threshold $\epsilon$ in Equation~\ref{l1min}, or equivalently designate a sparsity level $s$ so that $\|\beta\|_{0} \leq s$. As $\epsilon$ is difficult to determine for real data, in this paper we choose to set $s$ rather than $\epsilon$: this allows us to better compare the vertex classification performance through-out different sparsity levels, and we will argue that SRC is robust against $s$ in the next subsection and also the numerical experiments. Note that the constraint in Equation~\ref{l1min} can be replaced by $\phi = D \beta$ in a noiseless setting, but usually some parameters like $\epsilon$ or $s$ is required to achieve a parsimonious model, when dealing with high-dimensional or noisy data.

Although the SRC algorithm can always be used for supervised learning, it does not always perform well for arbitrary data sets; and it is necessary to understand why SRC is applicable to graphs. In \cite{wright2009robust}, it is argued that the face images of different classes lie on different subspaces, so that $\ell 1$ minimization is able to select training data of the correct class (i.e., the true but unknown class of the testing observation). Based on this subspace assumption, \cite{ElhamifarVidal2013} derives a theoretical condition for $\ell 1$ minimization to do perfect variable selection in sparse representation, i.e., all selected training data are from the correct class. This validates that sparse representation is a valuable tool with $\ell 1$ minimization under the subspace assumption. However, the subspace assumption requires an intrinsic low-dimensional structure for each class, which may not be satisfied for high-dimensional real data such as the adjacency matrix.

Furthermore, the motivation behind the popularity of $\ell 1$ minimization is its equivalence to $\ell 0$ minimization under certain conditions, such as the incoherence condition or restricted isometry property, see \cite{donoho2006most}, \cite{donoho2003optimally}, \cite{gribonval2003sparse}, \cite{candes2005decoding}, \cite{elad2002generalized}, \cite{rubinstein2010dictionaries}. But those conditions are often violated in the SRC framework, because the sample training data are usually correlated; and SRC does not necessarily need a unique or most sparse $\beta$ in order to do correct classification. As long as the sparse representation $\beta$ assigns dominating coefficients to data of the correct class, SRC can classify correctly.

Shen et al.\ \cite{ShenChenPriebe2015} proves SRC performance guarantee under a principal angle condition, which is similar to the condition in \cite{ElhamifarVidal2013}, but does not rely on the subspace assumption and does not require a unique and most sparse solution. The condition is easy to check for a given model and intuitive to understand: as long as the within-class principal angle is smaller than the between-class principal angle, $\ell 1$ minimization and OMP are able to assign dominating regression coefficients to training data of the correct class, so that SRC can perform well. Based on this direction, in the next subsection we derive a condition on the stochastic blockmodels so that the principal angle condition is satisfied, consequently achieving SRC consistency for SBM.

\subsection{SRC Consistency for SBM}
\label{SRCSection2}
Here we prove a consistency theorem for sparse representation classifier for vertex classification on the stochastic blockmodel, which provides theoretical performance guarantees of our proposed robust vertex classification. All proofs are put into the supplementary material.

For this subsection only, we first define for each $q=1, \ldots K$, 
\begin{equation}
\label{auxil}
Q_{q} \sim \sum_{k=1}^{K} \mathbf{1}_{\{Y=k\}} B_{kq},
\end{equation}
where $Y$ is the class label, $\mathbf{1}_{\{Y=k\}}$ is the indicator function with probability $\pi_{k}$, and $B_{kq}$ corresponds to the entry of the probability matrix $B$ generating SBM. Note that $\{Q_{q}\}$ and all their moments only depend on the prior probability $\pi$ and the block probability $B$.

Next we define the un-centered correlation as
\begin{align*}
\rho_{qr} = \frac{E(Q_{q}Q_{r})}{\sqrt{E(Q_{q}^{2})E(Q_{r}^{2})}},
\end{align*}
for each $1 \leq q \neq r \leq K$. Clearly $0 \leq \rho_{qr} = \rho_{rq} \leq 1$. 
 
Our first lemma proves a necessary and sufficient condition on the SBM parameters for adjacency columns of the same class to be asymptotically most correlated.

\begin{lemma}
\label{SBMLemma1}
Under the stochastic blockmodel, for an adjacency column of class $q$, its asymptotic most correlated column is of the same class $q$, if and only if the prior probability $\pi$ and the block probability matrix $B$ satisfy the following inequality:
\begin{equation}
\label{mainCondition}
\rho_{qr}^{2} \cdot \frac{E(Q_{r}^{2})}{E(Q_{q}^{2})} < \frac{E(Q_{r})}{E(Q_{q})}
\end{equation}
for all $r \neq q$.
\end{lemma}

When Lemma~\ref{SBMLemma1} holds for all $q$, it in fact guarantees that SRC at $s=1$ (or equivalently $1$-nearest-neighbor based on principal angle) is a consistent classifier for the stochastic blockmodel. To prove SRC consistency at any $s$, we need a second lemma.

\begin{lemma}
\label{SBMLemma2}
Denote $A_{(s)}$ as an $s \times n$ random matrix consisting of $s$ adjacency columns, and denote $C$ as a scalar vector of length $s$.

Suppose Equation~\ref{mainCondition} holds for the stochastic blockmodel. Then for any adjacency column $\alpha$ of class $q$, its within-class correlation (i.e., the correlation between $\alpha$ and another adjacency column of class $q$) is asymptotically larger than the correlation between $\alpha$ and $C \cdot A_{(s)}$, for any $A_{(s)}$ whose columns are not from class $q$ and any vector $C$ with non-negative entries. 

The above holds for any $s \geq 1$.
\end{lemma}

The above two lemmas essentially establish the principal angle condition in \cite{ShenChenPriebe2015}. They can guarantee that $\beta$ assigns dominating coefficients to training data of the correct class, which leads to SRC consistency for SBM.

\begin{theorem}
\label{SBMTheorem}
Suppose Equation~\ref{mainCondition} holds for the corresponding stochastic blockmodel for all $q \in [1,\ldots, K]$, and the sparse representation $\beta$ is constrained to be non-negative.

Then SRC is a consistent classifier for vertex classification of SBM, with $L_{n} \rightarrow 0$ as $n \rightarrow 0$. This holds for SRC implemented by either exact $\ell 1$ minimization or orthogonal matching pursuit at any $s \geq 1$.
\end{theorem}

Let us make some remarks regarding the theorem and its implication. Firstly, if the block columns are very close in their $\ell 1$ and $\ell 2$ norms in the measure space with respect to $\pi$ (i.e., $\frac{E(Q_{r})}{E(Q_{q})} \approx \frac{E(Q_{r}^{2})}{E(Q_{q}^{2})}$), then the theorem is very likely to hold for all $\rho_{qr} < 1$ and SRC is expected to perform well; if not, the block columns cannot be too highly correlated in order for the inequality to hold and for SRC to work; and if block $r$ is a scalar multiple of block $q$, the condition always fails and SRC cannot separate those two classes. In any case, if the adjacency matrix can be modeled by SBM, then it is very easy to estimate the model parameters and check Equation~\ref{mainCondition}. 

Secondly, even though Equation~\ref{mainCondition} is only sufficient and not necessary for SBM consistency at $s>1$, it is often the case that SRC is no longer consistent when Equation~\ref{mainCondition} is violated. Because when Equation~\ref{mainCondition} is violated for some $r$, the adjacency column of class $q$ is asymptotically most correlated with a column from class $r$, which usually causes SRC to misbehave. 

Thirdly, the theorem requires the sparse representation to be non-negative, which can be easily achieved in $\ell 1$ minimization; and \cite{BrucksteinEladZibulevsky2008}, \cite{Meinshausen2013}, \cite{SlawskiHein2013} show that eliminating the negative entries of $\beta$ has very nice theoretical properties in non-negative OMP and non-negative least square. Even though we do not explicitly use non-negative $\ell 1$ minimization or bound the coefficients, in our numerical experiments the negative entries of $\beta$ are almost never large, and $L_{n}$ clearly converges to $0$ for the SBM simulation in the numerical section. 

Fourthly, since the consistency result holds for SRC at any $s \geq 1$, we expect SRC to be robust in the choice of $s$, compared to the model selection of $\hat d$ for ASE procedures. This is demonstrated empirically in Section \ref{sec:experiments}. In particular, the two contamination scenarios essentially double the number of blocks comparing to the uncontaminated SBM; this causes the classification error of ASE to be no longer consistent unless the embedding dimension $d$ is adjusted accordingly, but SRC may remain consistent as long as the contaminated blocks still satisfy Equation~\ref{mainCondition}.

Lastly, we should note that even though the consistency results hold at any $s \geq 1$, in most experiments moderate $s$ helps the finite-sample performance comparing to $s=1$ or large $s$: One explanation is that the classifier itself is designed to favor a more parsimonious model as argued in \cite{wright2009robust}. Another explanation based on the consistency proof of \cite{ShenChenPriebe2015}, is that the sub-matrix of $D$ corresponding to the nonzero entries of $\beta$ should be full rank; this is always true when using $\ell 1$ minimization and OMP, but large $s$ may make the sub-matrix close to rank deficient (i.e., having singular values close to zero). Indeed in the numerical section, we will see that as long as the sparsity level $s$ is not too large relative to the sample size $n$, SRC can perform well; in addition, choosing smaller sparsity level has less computational cost. 

\section{Numerical Experiments}\label{sec:experiments} 
If the true model dimension is unknown, ASE$_{\hat{d}}$ may not be consistent. In particular, when contamination results in a changed model dimension, or the model dimension cannot be correctly estimated, the performance of subsequent classifiers may suffer. We consider a classifier robust if it can maintain relatively low misclassification rate under data contamination. Our sparse representation classifier (SRC) for vertex classification does not rely on the knowledge of the model dimension, is robust to the choices of sparsity level $s$, and achieves consistency with respect to all sparsity levels. 

Throughout this section, we use the orthogonal matching pursuit (OMP) to solve the $\ell 1$ minimization. In the following experiments, SRC$_s$ denotes the performance of SRC with varying sparsity levels $s$, and SRC$_5$ means $s=5$ by default. We use leave-one-out cross validation to estimate the classification error. The standard errors are small compared to the differences in performance. Our simulation experiments and real data analysis demonstrate that SRC for vertex classification performs well under varying sparsity levels, possesses higher robustness to contamination than 1NN$\circ$ASE$_{\hat{d}}$ and LDA$\circ$ASE$_{\hat{d}}$, and is an excellent tool for real data inference.

\subsection{Simulation}
We compare the robustness of SRC with two vertex classifiers: 1NN$\circ$ASE$_{\hat{d}}$ and LDA$\circ$ASE$_{\hat{d}}$, both of which achieve the asymptotic Bayes error when $\hat{d} =d $ with no contamination, and $\hat{d} = d_{\text{occ}}$ or $d_{\text{rev}}$ in the contamination model. 

\subsubsection{No Contamination}
We simulate the probability matrix for an uncontaminated stochastic blockmodel $\mathbb{G}_{\text{un}}$ with $K=2$ blocks ($Y \in \{1,2\}$) and parameters
\[
B_{\text{un}} =
\left( {\begin{array}{cc}
 0.7 & 0.32 \\
 0.32 & 0.75
 \end{array} } \right)
\]
\begin{equation}
\pi_{\text{un}}=[0.4, \enspace 0.6]^{T}.
\label{eq:simParams}
\end{equation}
The SBM parameters in Equation \ref{eq:simParams} in fact satisfies the theoretical condition in Equation~\ref{mainCondition}, so we expect SRC to perform well in this case.

\begin{figure*}
  \centering
    \includegraphics[width = 3in]{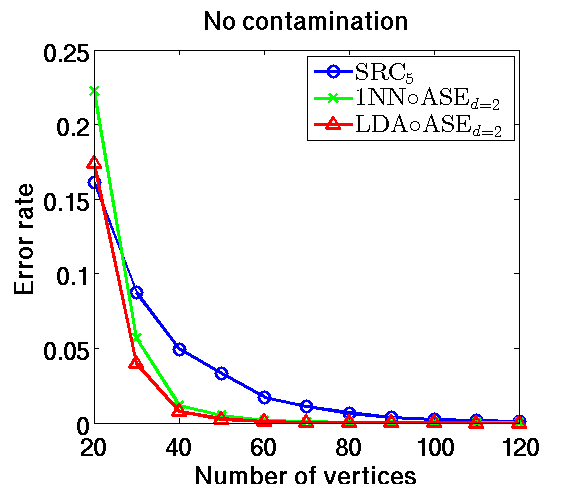}
	\includegraphics[width = 3in]{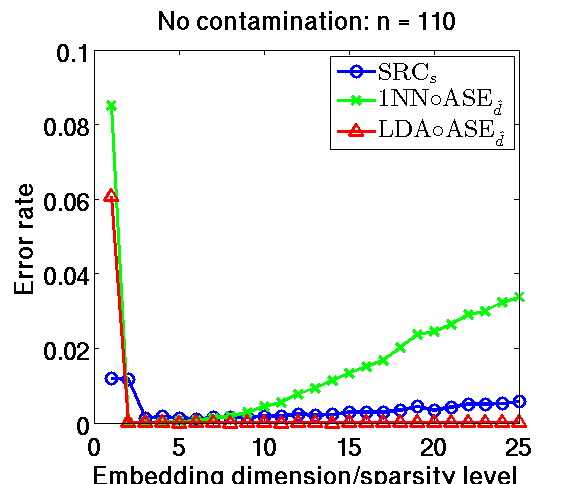}
    \caption{\textbf{Classification performance under no contamination.} We simulate $100$ SBMs with $B_{\text{un}}$, $\pi_{\text{un}}$ given in Equation \ref{eq:simParams}, and show the average the misclassification error over the $100$ Monte Carlo replicates. (Left): When the true model dimension $d = 2$ is known, SRC does not outperform 1NN$\circ$ASE$_{d = 2}$ or LDA$\circ$ASE$_{d = 2}$ for $n \in [20, 120]$. (Right): Do the same vertex classification using various $s, \hat{d}$ at $n = 110$.  }
    \label{fig:clean_data}
\end{figure*}

We first assess the performance of all classifiers in the uncontaminated model, assuming the true model dimension $d = 2$ is known. As seen in the left plot of Figure \ref{fig:clean_data}, LDA$\circ$ASE$_{d = 2}$ performs the best for all $n \in \{30, 40, \ldots, 120\}$. In this ideal setting, SRC does not outperform 1NN$\circ$ASE$_{d = 2}$ or LDA$\circ$ASE$_{d = 2}$, but all classifiers converge to $0$ error as $n$ increases, as expected based on our theoretical derivation.

Then we fix the number of vertices $n =110$ and vary the sparsity level $s$ and embedding dimension $\hat{d}$. The right plot of Figure \ref{fig:clean_data} exhibits the three classifiers' performance. SRC$_s$ performs well throughout $s$, so does LDA$\circ$ASE$_{\hat{d}}$ except at $\hat{d} = 1$, while 1NN$\circ$ASE$_{\hat{d}}$ degrades significantly with increasing $\hat{d}$ or $\hat{d}=1$.


\subsubsection{Under Contamination}
Now we assess the robustness of SRC, 1NN$\circ$ASE$_{\hat{d}}$ and LDA$\circ$ASE$_{\hat{d}}$ under contamination using the same parameter setting as Equation~\ref{eq:simParams}. If the model dimension $d = 2$ is known and the exact contamination is known, then $\hat{d} = 4$ is best for subsequent classification of the contaminated data; otherwise $\hat{d}$ will be set to $2$, either due to not knowing the contamination or due to estimating $\hat{d}$ by the profile likelihood procedure in \cite{zhu2006automatic}, as seen in Figure \ref{fig:ase_scree1} and Figure \ref{fig:ase_scree2}.   

\begin{figure*}
  \centering
    \includegraphics[width=2.in]{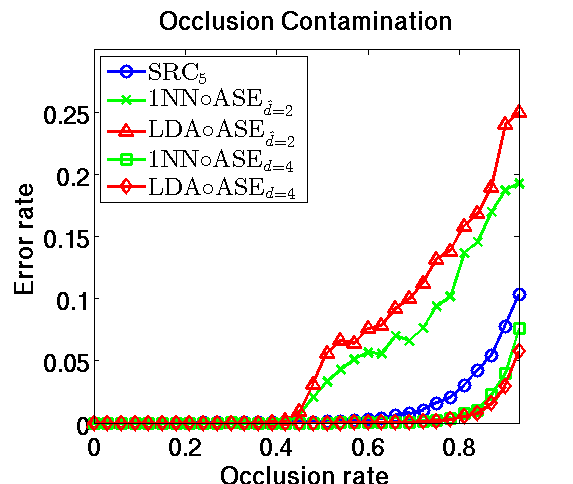}
        \includegraphics[width=2.in]{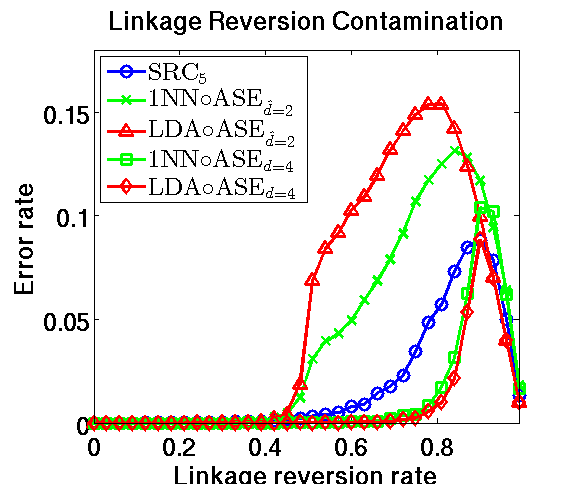}
        \includegraphics[width=2.in]{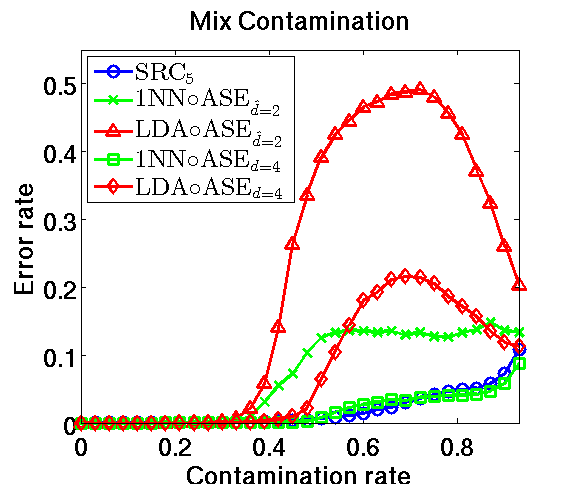}
    \caption{\textbf{Classification performance under three types of contamination.} We simulate $100$ SBMs with $B_{\text{un}}$, $\pi_{\text{un}}$ given in Eq.\ \ref{eq:simParams}, set $n=200$, contaminate the data accordingly, and present the average misclassification error for the five classifiers over the $100$ Monte Carlo replicates. SRC at $s=5$ exhibits robust performance compared to 1NN$\circ$ASE$_{\hat{d} = 2}$ and LDA$\circ$ASE$_{\hat{d} = 2}$, throughout all type of contamination with varying contamination rates.}
    \label{fig:err_occlusion}
\end{figure*}

Figure \ref{fig:err_occlusion} presents the misclassification error of SRC, 1NN$\circ$ASE$_{\hat{d}}$ and LDA$\circ$ASE$_{\hat{d}}$ under occlusion contamination, linkage reversion contamination, and a mixed combination of both contamination, for $s=5$ and $\hat{d} = 2, 4$ respectively. The x-axis stands for the contamination rate, while the y-axis stands for the classification error. In case of occlusion, all classifiers degrade as the contamination rate increases, due to less density in the graph. And in case of linkage reversion, all classifiers degrade first due to a weaker block signal, and then improve when the contamination rate increases above $0.95$, because the reversed block signal becomes stronger. As to the mix contamination, it is done as follows: first, we randomly select $100p\%$ vertices and occlude their connectivity; secondly, we randomly select $100p\%$ vertices (some may have already been occluded) and reverse their connectivity. In this scenario, the degradation in classification performance comes from both occlusion and linkage reversion contamination. 

For both occlusion and linkage reversion, LDA$\circ$ASE$_{d = 4}$ is the best classifier, followed by 1NN$\circ$ASE$_{d = 4}$. SRC is slightly inferior, but is significantly better than LDA$\circ$ASE$_{\hat{d} = 2}$ and 1NN$\circ$ASE$_{\hat{d} = 2}$. For the mixed contamination, SRC and 1NN$\circ$ASE$_{d = 4}$ are the best classifiers, which perform much better than the others. This indicates that SRC is robust against the contamination, while subsequent classification after spectral embedding may suffer from model dimension misspecification and data contamination.

\begin{figure*}
\centering
    \includegraphics[width=1.5in]{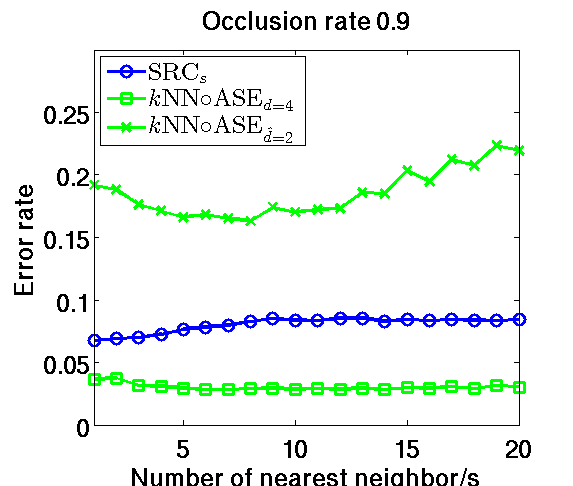}
    \includegraphics[width=1.5in]{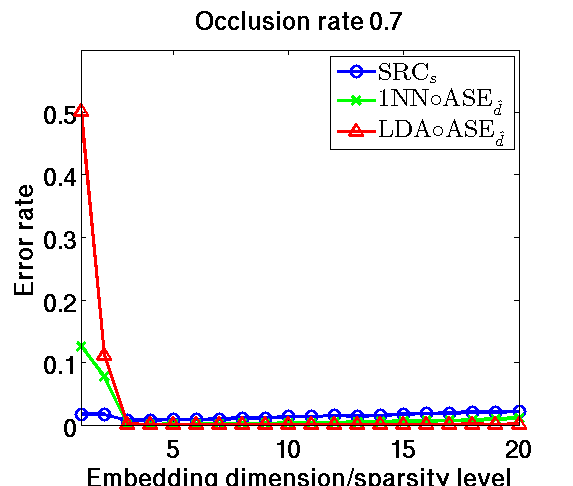}
    \includegraphics[width=1.5in]{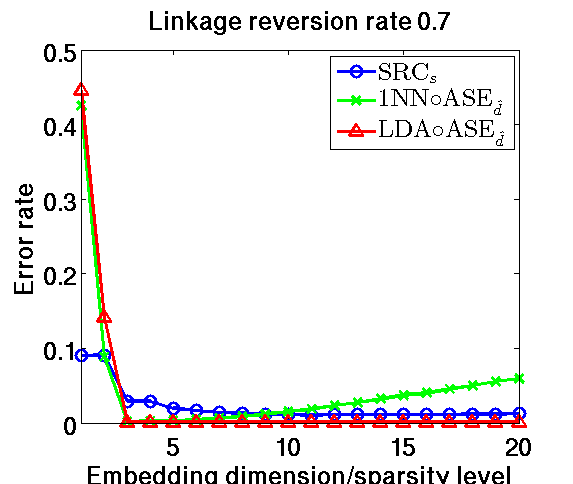}
    \includegraphics[width=1.5in]{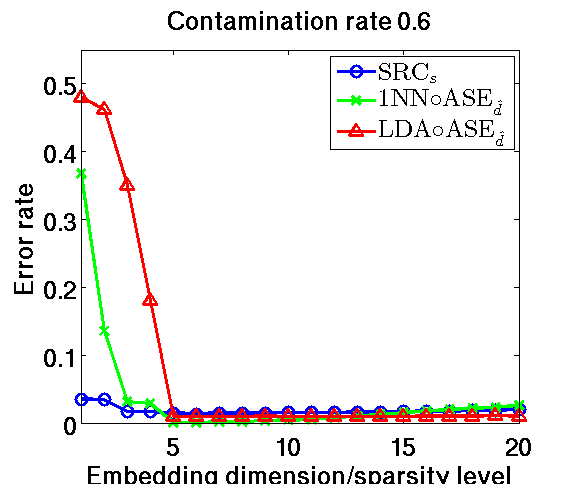}
\caption{Under the same setting of Figure~\ref{fig:err_occlusion}, the first plot varies the choice of neighborhood $k$ in kNN$\circ$ASE, and compare with SRC$_s$ with varying $s$. The other three plots compare the classification error of SRC$_s$, 1NN$\circ$ASE$_{\hat{d}}$, and LDA$\circ$ASE$_{\hat{d}}$ throughout $s=\hat{d} \in [1,\ldots,20]$. SRC exhibits stable performance with respect to the sparsity level $s$. }
\label{fig:err_flipping}
\end{figure*}

Note that SRC also has a model selection parameter, namely the sparsity level $s$. Thus in Figure~\ref{fig:err_flipping} we plot SRC error with respect to the sparsity level $s \in [1,\ldots,20]$, as well as LDA$\circ$ASE$_{d}$ and 1NN$\circ$ASE$_{d}$ with respect to the embedding dimension $d \in [1,\ldots,20]$. Furthermore, because we have fixed the number of nearest neighbor to be $1$ so far, the first plot in Figure~\ref{fig:err_flipping} is used to show that varying the number of nearest-neighbor does not help kNN$\circ$ASE$_{\hat{d}=2}$ for $k \in [1,\ldots,20]$. All plots in Figure~\ref{fig:err_flipping} show that SRC is stable with respect to the sparsity level $s$, while ASE methods are less robust with respect to the dimension choice.

\subsection{Real Data Experiments}
We apply SRC to several real datasets. We binarize, symmetrize the adjacency matrix and set the diagonals to be zero. We followed \cite{marchette2011vertex} and \cite{scheinerman2010modeling}, which suggest imputing the diagonal of the adjacency matrix to improve performance. We vary the embedding dimension $\hat{d} $ for 1NN$\circ$ASE$_{\hat{d}}$ and LDA$\circ$ASE$_{\hat{d}}$, and the sparsity level $s$ for SRC$_s$.

\subsubsection{\textit{C.elegans} Neural Connectome}
\begin{figure}
\centering
\includegraphics[width = 3in]{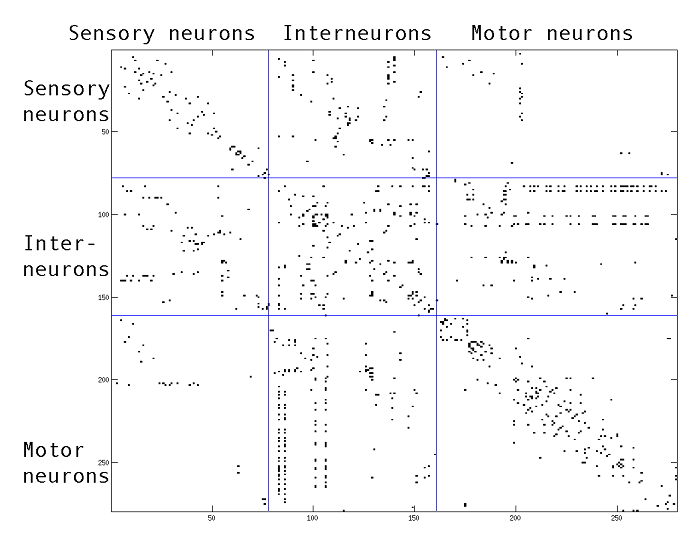}
\includegraphics[width = 3in]{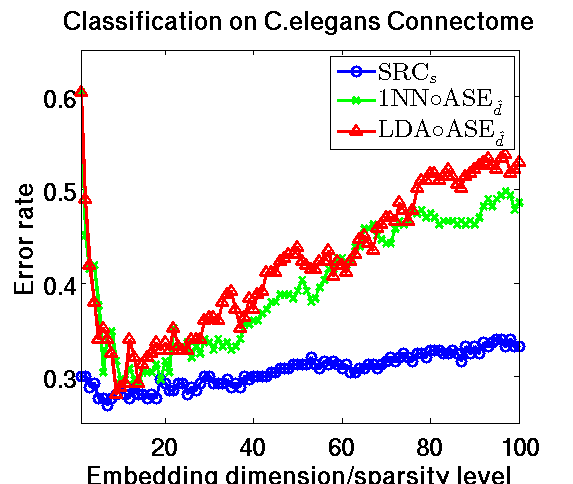}
\caption[Vertex classification performance on the \textit{C.elegans} network]{(Top): The adjacency matrix of the \textit{C.elegans} neural connectome is sorted according to the classes of the neurons. A three-block structure is exhibited. (Bottom): Vertex classification performance on the \textit{C.elegans} network. As we vary the sparsity level $s$ and embedding dimension $\hat{d}$, SRC$_s$ demonstrates superior and stable performance compared to 1NN$\circ$ASE$_{\hat{d}}$ and LDA$\circ$ASE$_{\hat{d}}$.}
\label{fig:err_celegans}
\end{figure}

We apply SRC to the electric neural connectome of Caenorhabditis elegans (\textit{C.elegans}) \cite{hall1991posterior}, \cite{goldschmidt1908nervensystem}, \cite{chen2015celegans}. The hermaphrodite \textit{C.elegans} somatic nervous system has $279$ neurons (\cite{varshney2011structural}). Those neurons are classified into $3$ classes: motor neurons ($42.29 \%$), interneurons ($29.75 \%$) and sensory neurons ($ 27.96 \%$). The adjacency matrix is seen in the top of Figure \ref{fig:err_celegans}. The graph has density $\frac{514}{{279 \choose 2}} = 1.32\%$. The objective is to predict the classes of the neurons, and the chance line for this classification task is $57.71 \%$.

The bottom of Figure \ref{fig:err_celegans} demonstrates the performance of the three classifiers. Both LDA$\circ$ASE$_{\hat{d}}$ and 1NN$\circ$ASE$_{\hat{d}}$ improve in performance as $\hat{d}$ increases to $10$, since more signal is included in the embedded space; and as $\hat{d}$ continues to increase to $100$, both classifiers gradually degrade in performance, since more noise is included. The exhibited phenomenon is due to bias-variance trade-off. In comparison, SRC$_s$ has stable performance with respect to the sparsity level $s \in [1,\ldots,100]$, which outperforms LDA$\circ$ASE$_{\hat{d}}$ and 1NN$\circ$ASE$_{\hat{d}}$. This demonstrates that SRC is a practical tool in random graph inference.

\subsubsection{Adjective and Noun Network}
\begin{figure}
\centering
\includegraphics[width = 3in]{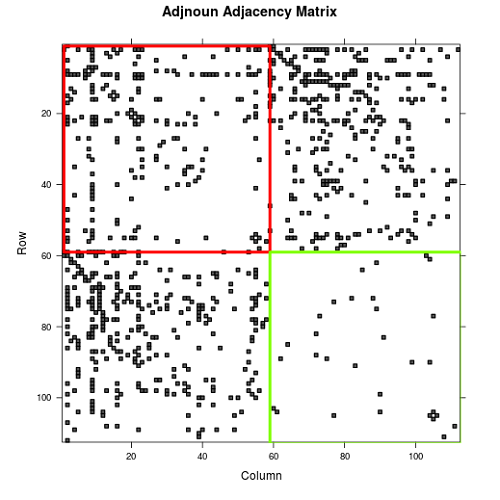}
\includegraphics[width = 3in]{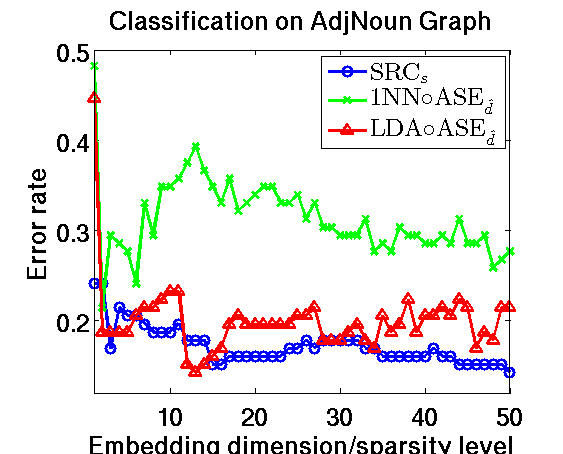}
\caption[Vertex classification performance on the adjective and noun network]{(Top): Adjacency matrix of adjective noun network, where each class is more likely to communicate with the other class than itself. (Bottom): Vertex classification performance on the adjective and noun network. SRC$_s$ demonstrates robust performance compared to 1NN$\circ$ASE$_{\hat{d}}$, and LDA$\circ$ASE$_{\hat{d}}$.}
\label{fig:err_adjnoun}
\end{figure}

The AdjNoun graph, collected in \cite{newman2006finding}, is a network containing frequently used adjectives and nouns from the novel ``David Copperfield" by Charles Dickens. The vertices are the 60 most frequently used adjectives and 60 most frequently used nouns in the book. The edges are present if any pair of words occur in an adjacent position in the book. The chance error is $48.21\%$. The adjacency matrix of the adjective noun network suggests that the connectivity between nouns and adjectives are more frequent than the connectivities among nouns and the connectivities among adjectives respectively, as seen in the top of Figure \ref{fig:err_adjnoun}. 

We apply SRC$_s$, 1NN$\circ$ASE$_{\hat{d}}$, and LDA$\circ$ASE$_{\hat{d}}$ on this dataset, and vary the embedding dimension $\hat d \in \{1,2,\ldots, 50\}$ and the sparsity level $s \in \{1,2,\ldots, 50\}$. Performance of the three classifiers is seen in the bottom of Figure \ref{fig:err_adjnoun}. SRC$_s$ again exhibits stable performance with respect to various sparsity level $s$, comparing to 1NN$\circ$ASE$_{\hat{d}}$ and LDA$\circ$ASE$_{\hat{d}}$. Note that as the number of vertices is only $120$, we limit the sparsity level to $50$ in this experiment.

\subsubsection{Political Blog Sphere}
\begin{figure}
\centering
\includegraphics[width=3in]{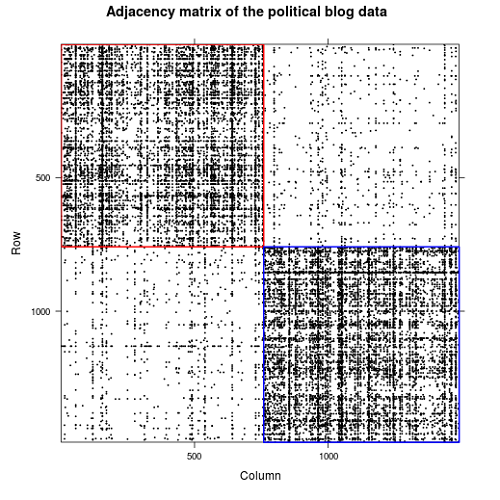}
\includegraphics[width=3in]{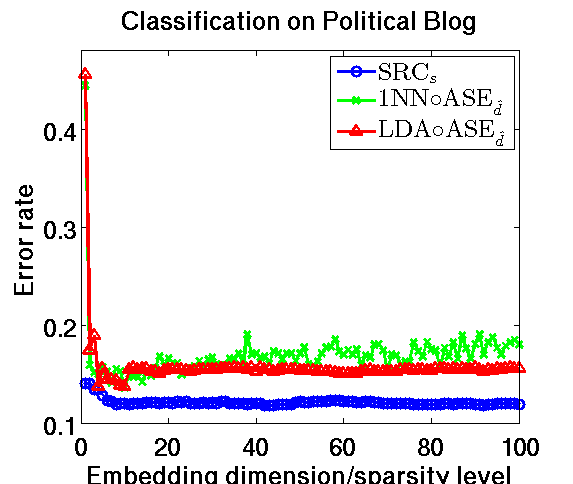}
\caption[Vertex classification performance on the political blog network]{(Top): Adjacency matrix of political blog sphere, exhibiting strong connectivity within class. (Bottom): Vertex classification performance on the political blog network. SRC$_s$ demonstrates superior and very stable performance with respect to various sparsity levels $s$, comparing to 1NN$\circ$ASE$_{\hat{d}}$ and LDA$\circ$ASE$_{\hat{d}}$.}
\label{fig:src_blog}
\end{figure}

The political blog sphere was collected in February 2005 \cite{adamic2005political}. The vertices are blogs during the time of the 2004 presidential election, and edges exist if the blogs are linked. The blogs are either liberal or conservative, which sum up to $n=1490$ vertices. The top of Figure \ref{fig:src_blog} demonstrates the adjacency matrix of the blog network, which reflects a strong two-block signal. 

The performance of three classifiers is shown in the bottom of Figure \ref{fig:src_blog}, with varying sparsity level $s$ and dimension choice $\hat{d}$ up to $100$. SRC$_s$ has very stable and superior performance with respect to various sparsity level, and always outperforms 1NN$\circ$ASE$_{\hat{d}}$ and LDA$\circ$ASE$_{\hat{d}}$. It is worthwhile to point out that this dataset can be modeled by SBM as shown in \cite{OlhedeWolfe2014}; and the sparsity limit $100$ is relatively small comparing to the number of vertices $n$ here, which is the reason why SRC$_s$ is very stable up to $s=100$.

\subsubsection{Political Book Graph}
\begin{figure}
\centering
\includegraphics[width=3in]{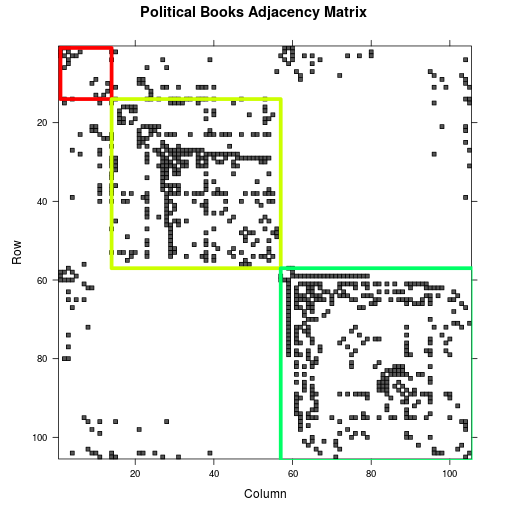}
\includegraphics[width=3in]{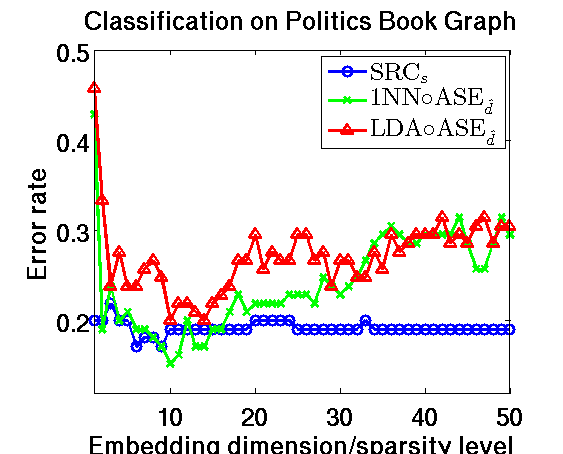}
\caption[Vertex classification on the political book graph]{(Top): Adjacency matrix of political book graph. (Bottom): Classification performance on the political book graph.}
\label{fig:polibook}
\end{figure}

The political book graph contains 105 books about US politics and sold by Amazon.com \cite{newman2006finding}. The edges exist if any pairs of books were purchased by the same customer. There are 3 class labels on the books: liberal (46.67$\%$), neural (40.95$\%$) and conservative (12.28$\%$). The adjacency matrix of this dataset and the performance of the three classifiers are seen in the top of Figure \ref{fig:polibook}.

The bottom of Figure \ref{fig:polibook} shows that SRC$_s$ is very stable with respect to the sparsity level, and usually better than 1NN$\circ$ASE$_{\hat{d}}$ and LDA$\circ$ASE$_{\hat{d}}$, but the optimal error is achieved by 1NN$\circ$ASE$_{\hat{d}=10}$.

\section{Discussion}\label{discussion}
Adjacency spectral embedding is a feature extraction approach for latent position graphs. When feature extraction is composed with common classifiers such as nearest-neighbor or discriminant analysis, the choice of feature space or embedding dimension is crucial. Given the model dimension $d$ for a stochastic blockmodel, ASE$_{d}$ is consistent and the subsequent vertex classification via 1NN$\circ$ASE$_{d}$ or LDA$\circ$ASE$_{d}$ is asymptotically Bayes optimal. And the success of ASE procedures clearly depends on the knowledge of $d$, as illustrated in the experiments.

However, in practical settings, the model dimension $d$ is usually unknown, and there may exist data contamination. In this paper, we present a robust vertex classifier via sparse representation for graph data. The sparse representation classifier does not need information of the model dimension, can achieve consistency under a mild condition for SBM parameters, and is robust against the choice of sparsity levels. As seen in the simulation studies using SBM, SRC may not outperform 1NN$\circ$ASE$_{d}$ and LDA$\circ$ASE$_{d}$ when $d$ is known, but does outperform 1NN$\circ$ASE$_{\hat{d}}$ and LDA$\circ$ASE$_{\hat{d}}$ where $\hat{d}$ is chosen using the scree plot of the adjacency matrix. In the real data experiments, most of the time SRC outperforms 1NN$\circ$ASE$_{\hat{d}}$ and LDA$\circ$ASE$_{\hat{d}}$ for varying $\hat{d}$, and is very stable with respect to the sparsity level $s$. The numerical studies strongly indicate that SRC is a valuable tool for random graph inference. 

For SRC implementation, we only considered orthogonal matching pursuit (OMP) to solve the $\ell 1$ minimization problem. Different implementations of $\ell 1$ minimization are explored in \cite{ShenChenPriebe2015}, and using a different algorithm may yield slightly different classification performance for SRC. 

\begin{figure}[!ht]
  \centering
    \includegraphics[width=3.3in]{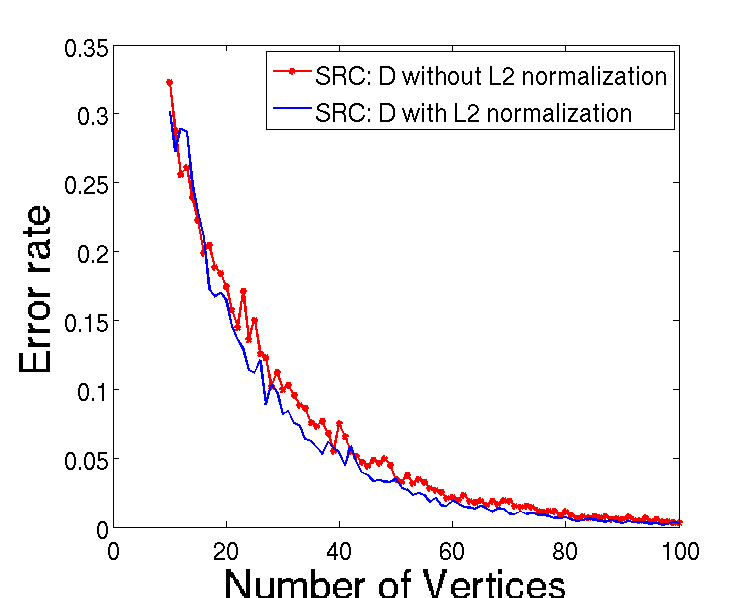}
    \caption{\textbf{Examination of SRC performance with or without $\ell 2$ normalization on columns of $D$.} We compare SRC performance when columns of $D$ are $\ell 2$ normalized and when columns of $D$ are not $\ell 2$ normalized. The parameters $B$ and $\pi$ are given in Eq.\ \ref{eq:simParams} with $n \in \{10,\ldots, 100\}$ and we run $100$ Monte Carlo replicates for each $n$. We see an improvement in SRC performance when $\ell 2$ normalization is applied. The Wilcoxon signed rank test reports a $p$-value less than $0.05$ under the the null hypothesis that the error difference SRC$_{\text{error}, \ell 2 }-$ SRC$_{\text{error}, \text{no }\ell 2}$ comes from a distribution with zero median. }
    \label{fig:L2orNotL2}
\end{figure}

Another interesting question is the effect of normalization, namely the transformation of $A$ into $D$ in Algorithm 1. The normalization effect is usually difficult to quantify; but empirically, we see improvement in SRC performance under $\ell 2$ normalization, as illustrated in Figure \ref{fig:L2orNotL2}. Note that the SBM parameters satisfy the condition in Equation~\ref{mainCondition}, so we expect SRC to perform well in the normalized case; furthermore, in the figure SRC error is very close to $0$ as $n$ increases, despite the fact that the non-negative constraint is not used in the algorithm (which is used in the consistency proof).

\section{Appendix} \label{sec:appendix}

\subsection{The Eigen-structure of the Occlusion Model}
An event occurs ``almost always", if with probability 1, the event occurs for all but finitely many $n$. 
\begin{proposition}
It always holds that $\sigma_{1}(P_{occ}) \leq \sigma_{1}(P_{\text{un}}) \leq n$.
\label{thm:sigma1_p_occ_less_n}
\end{proposition}

\begin{proof}
Suppose the set of the contaminated vertices is $\mathcal{I} := \{i_1, i_2,\ldots, i_l \}$. Let $P_s'$ denote the principal submatrix of $P_{\text{un}} \in \mathbb{R}^{|\mathcal{I}| \times |\mathcal{I}|}$ obtained by deleting the $V\setminus\mathcal{I}$ columns and the corresponding $V\setminus\mathcal{I}$ rows. $P_s'$ is symmetric.

Note that $P_{\text{un}} = P_{\text{occ}} + P_{s}$, where $P_s$ is symmetric, $ P_{s} = P_s'$ at $\{i_1, i_2,\ldots, i_l \}$-th columns and  $\{i_1, i_2,\ldots, i_l \}$-th rows, and $P_s = 0$ everywhere else. 
By Weyl's Theorem \cite{horn2012matrix}, $\sigma_1(P_{\text{occ}}) + \min_{\sigma} \sigma(P_s) \leq \sigma_1(P_{\text{occ}} + P_{s}) = \sigma_1(P_{\text{un}})$. Thus, $\sigma_{1}(P_{\text{occ}}) \leq \sigma_{1}(P_{\text{un}})$.

Since $P_{\text{un}} \in [0, 1]^{n}$, $P_{\text{un}}P_{\text{un}}^{T} = P_{\text{un}}P_{\text{un}}$ is a non-negative and symmetric matrix with entries bounded by $n$. Then each row sum is bounded by $n^{2}$. Thus, 
$\sigma_{1}^{2}(P_{\text{un}}) = \sigma_{1}(P_{\text{un}}^2) = \sigma(P_{\text{un}}P_{\text{un}}^{T}) \leq n^2$, giving 
$\sigma_{1}(P_{\text{un}}) \leq n.$

\end{proof}

\begin{proposition}
It always holds that $\sigma_{2d+1}(P_{occ}) =0$. It almost always holds that $\sigma_{2d}(P_{occ}) \geq \min(p_{0}, 1-p_{0})\alpha\gamma n$. rank($P_{occ}) = 2d$.
\label{thm:thm_2}
\end{proposition}

\begin{proof}
The Guassian elimination of $B_{\text{occ}}$ is given by 
\begin{equation}
B_{\text{occ}} \sim
\left( {\begin{array}{cc}
 B_{\text{un}} & B_{\text{un}} \\
 0_{K \times K} & B_{\text{un}}
 \end{array} } \right). 
\end{equation}
Since rank$(B_{\text{un}})  = d$, rank$(B_{\text{occ}}) = 2d$. Then there exist $\mu = \left( {\begin{array}{cc}
 \nu &  0_{K \times K}\\
 \nu & -\nu
 \end{array} } \right) \in \mathbb{R}^{2K \times 2d}$ and $\tilde{\mu} =  \left( {\begin{array}{cc}
 \nu &  0_{K \times K}\\
 \nu & \nu
 \end{array} } \right)\in \mathbb{R}^{2K \times 2d}$ such that $B_{\text{occ}} = \mu \tilde{\mu}^{T}$. Let $\mathcal{X}_{\text{occ}} \in \mathbb{R}^{n \times 2d}$ and $\tilde{\mathcal{X}}_{\text{occ}} \in \mathbb{R}^{n \times 2d}$ with row $u$ given by $\tilde{\mathcal{X}}_{\text{occ}, u} = \tilde{\mu}_{Y_{u}}$. By the parametrization of SBM as RDPG model, $P_{\text{occ}} = \mathcal{X}_{\text{occ}}\tilde{\mathcal{X}}_{\text{occ}}^{T}$. Since $\mathcal{X}_{\text{occ}}, \tilde{\mathcal{X}}_{\text{occ}}$ are at most rank $2d$, then $\sigma_{2d+1}(P_{\text{occ}}) =0$. 
 
Since the following holds:

\begin{eqnarray}
\mu \mu^{T} &=& \tilde{\mu}\tilde{\mu}^{T} = 
\left( {\begin{array}{cc}
 \nu\nu^{T} & \nu\nu^{T} \\
 \nu\nu^{T} & 2\nu\nu^{T}
 \end{array} } \right) \\
 &=& \left( {\begin{array}{cc} 
 \nu\nu^{T} & \nu\nu^{T} \\
 \nu\nu^{T} & \nu\nu^{T}
 \end{array} } \right) + \left( {\begin{array}{cc}
 0_{K \times K} & 0_{K \times K} \\ 
 0_{K \times K} & \nu\nu^{T}
 \end{array} } \right), \nonumber
\end{eqnarray}

by Weyl's theorem \cite{horn2012matrix},
\begin{eqnarray}
&\min & \lambda_{i}(\mu \mu^{T}) = \min\lambda_{i}(\tilde{\mu} \tilde{\mu}^{T}) \\
& \geq & \min\lambda_{i}\left( {\begin{array}{cc} 
 \nu\nu^{T} & \nu\nu^{T} \\
 \nu\nu^{T} & \nu\nu^{T}
 \end{array} } \right) + \min\lambda_{i}\left( {\begin{array}{cc}
 0_{K \times K} & 0_{K \times K} \\ 
 0_{K \times K} & \nu\nu^{T}
 \end{array} } \right) \nonumber \\ 
& \geq & \gamma + 0 = \gamma.
\end{eqnarray}
Moreover, we have
\begin{equation}
\min_{i \in [2K]}(\pi_{\text{occ}, i}) = \min(p_{o}\pi_{\text{un}, i}, (1-p_{o})\pi_{\text{un}, i}) \geq \min(p_{o}, 1-p_{o})\gamma.
\end{equation}

The eigenvalues of $P_{\text{occ}}P_{\text{occ}}^{T}$ are the same as the nonzero eigenvalues of $\tilde{\mathcal{X}}_{\text{occ}}^{T}\tilde{\mathcal{X}}
_{\text{occ}}\mathcal{X}_{\text{occ}}^{T}\mathcal{X}_{\text{occ}}$. it almost always holds that $n_{i} \geq \min(p_{o}, 1-p_{0})\gamma n$ for all $i \in [2K]$ so that 
\begin{eqnarray}
\mathcal{X}_{\text{occ}}^{T}\mathcal{X}_{\text{occ}} &=& \sum_{i=1}^{2K}n_{i}\mu_{i}\mu_{i}^{T} = \min(p_{o}, 1-p_{o})\gamma n \mu^{T} \mu \nonumber \\ 
&+& \sum_{i=1}^{2K}(n_{i} - \min(p_{o}, 1-p_{o})\gamma n) \mu_{i}\mu_{i}^{T}. 
\end{eqnarray}

The first term has $\min \lambda_{i}$ bounded below by $\alpha \min(p_{o}, 1 - p_{o}) \gamma$. This means $\lambda_{2d}(\mathcal{X}_{\text{occ}}^{T}\mathcal{X}_{\text{occ}}) \geq \alpha \min(p_{o}, 1 - p_{o}) \gamma$. For the exact same argument, $\lambda_{2d}(\tilde{\mathcal{X}}_{\text{occ}}^{T}\tilde{\mathcal{X}}_
{\text{occ}}) \geq \alpha \min(p_{o}, 1 - p_{o}) \gamma$. $\tilde{\mathcal{X}}_{\text{occ}}^{T}\tilde{\mathcal{X}}_
{\text{occ}}\mathcal{X}_{\text{occ}}^{T}\mathcal{X}_{\text{occ}}$ is the product of two positive semi-definite matrices. Then,
\begin{eqnarray}
\lambda_{2d}(\tilde{\mathcal{X}}_{\text{occ}}^{T}\tilde{\mathcal{X}}_{\text{occ}}Z_{\text{occ}}^{T}Z_{\text{occ}}) &\geq & \lambda_{2d}(\tilde{\mathcal{X}}_{\text{occ}}^{T}\tilde{\mathcal{X}}_{\text{occ}})\lambda_{2d}(\mathcal{X}_{\text{occ}}^{T}\mathcal{X}_{\text{occ}}) \nonumber \\
&\geq& (\alpha\min(p_{0}, 1-p_{0})\gamma n)^2. \nonumber
\end{eqnarray}
This gives
\begin{equation}
\lambda_{2d}(\mathbf{P}_{\text{occ}}) \geq \alpha\min(p_{0}, 1-p_{0})\gamma n = \min(p_{0}, 1-p_{0})\alpha\gamma n.
\end{equation}
Since $\lambda_{2d}(P_{\text{occ}}) \geq  0$ almost always and $\sigma_{2d+1}(P_{\text{occ}}) =0$ always, then rank$(P_{\text{occ}}) = d$.
\end{proof}

\begin{proposition}
$B_{\text{occ}}$ has $d$ positive eigenvalues and $d$ negative eigenvalues.
\end{proposition}

\begin{proof}
Let the eigen-decomposition of $B_{\text{un}}$ given by  $\Xi \Psi \Xi^{T}$, where $\Xi \in \mathbb{R}^{K \times K}$ is orthogonal and $\Psi = \text{Diag}(\psi_{1}, ..., \psi_{k}) \in \mathbb{R}^{K \times K}$ is diagonal. We have the following congruent relation:
\begin{eqnarray}
\left(  {\begin{array}{cc}
 B_{\text{un}} & B_{\text{un}} \\
 B_{\text{un}} & 0_{K \times K}
 \end{array} } \right) & = & 
\left( {\begin{array}{cc}
 I_{K \times K} & 0_{K \times K} \\
 I_{K \times K} & I_{K \times K}
 \end{array} } \right)  \nonumber \\
 &\times & \left( {\begin{array}{cc}
 B_{\text{un}} & 0_{K \times K} \\
 0_{K \times K} & -B_{\text{un}}
 \end{array} } \right) \nonumber \\
 &\times & \left( {\begin{array}{cc}
 I_{K \times K} & 0_{K \times K} \\
 I_{K \times K} & I_{K \times K}
 \end{array} } \right)^{T} \nonumber  \\ 
 &=& \left({\begin{array}{cc}
 \Xi & 0_{K \times K} \\
 \Xi & \Xi
 \end{array} } \right) \nonumber \\
 &\times &\left( {\begin{array}{cc}
 \Psi & 0_{K \times K} \\
 0_{K \times K} & -\Psi
 \end{array} } \right) \nonumber \\
 &\times & \left( {\begin{array}{cc}
 \Xi & 0_{K \times K} \\
 \Xi & \Xi
 \end{array} } \right)^{T}.
\end{eqnarray}
Hence, $B_{\text{occ}}$ and $\left( {\begin{array}{cc}
 \Psi & 0_{K \times K} \\
 0_{K \times K} & -\Psi
 \end{array} } \right)$ are congruent. By Sylvester's law of Inertia \cite{horn2012matrix}, they have the same number of positive, negative and zero eigenvalues. $\Psi$ has $d$ positive diagonal entries since rank$(B_{\text{un}})=d$. Similarly, $-\Psi$ has $d$ negative diagonal entries. Hence, $B_{\text{occ}}$ has $d$ positive eigenvalues and $d$ negative eigenvalues.
\end{proof}

\begin{proposition}
Assuming $|\lambda_{1}(P_{\text{occ}})| \geq |\lambda_{2}(P_{\text{occ}})| \geq \ldots \geq |\lambda_{2d}(P_{\text{occ}})|$, then $|\{i: \lambda_{i}(P_{\text{occ}}) < 0 \}| = |\{i: \lambda_{i}(P_{\text{occ}}) < 0 \}| = d$. That is, the number of positive eigenvalues of $P_{occ}$ is the same as the number of negative eigenvalues of $P_{\text{occ}}$, and it equals $d$.
\end{proposition}

\begin{proof}
Let $Z \in \{0,1\}^{n \times 2K}$ denote the matrix, where each row $i$ is of the form $(0, \ldots, 1, 0, \ldots, 0)$, where 1 indicates the block membership of vertex $i$ in the occlusion stochastic blockmodel. Then $P_{\text{occ}} = ZB_{\text{occ}}Z^T$. Note that $P_{\text{occ}}$ has the same number of nonzero eigenvalues as $Z^T ZB_{\text{occ}}$. Let $D_Z := Z^T Z \in \mathbb{N}^{2K \times 2K}$ and note that $D_Z$ is a diagonal matrix with nonnegative diagonal entries, where each diagonal entry denotes the number of vertices belonging to block $k \in [K]$. With high probability, $D$ is positive definite, as the number of vertices in each block is positive. Then the number of nonzero eigenvalues of  $P_{\text{occ}}$ is the same as the number of nonzero eigenvalues of $Z^T ZB_{\text{occ}} = DB_{\text{occ}} = \sqrt{D_Z}\sqrt{D_Z}B_{\text{occ}} = \sqrt{D_Z}B_{\text{occ}}\sqrt{D_Z}$. By Sylvester's law of Inertia \cite{horn2012matrix}, the number of positive eigenvalues of $\sqrt{D_Z}B_{\text{occ}}\sqrt{D_Z}$ is the same as the number of positive eigenvalues of $B_{\text{occ}}$, and the number of negative eigenvalues of $\sqrt{D_Z}B_{\text{occ}}\sqrt{D_Z}$ is the same as the number of negative eigenvalues of $B_{\text{occ}}$, thus proving our claim.
\end{proof}

\subsection{SRC Consistency Proof}
\subsubsection{Proof of Lemma~\ref{SBMLemma1}}

\begin{proof}
We first prove that an adjacency column from class $q$ is asymptotically most correlated with another column of the same class, if and only if Equation~\ref{mainCondition} is satisfied.   

Suppose the first two vertices $1, 2$ are from class $1$, and vertices $3$ is of class $2$. Without loss of generality, let us prove that $A_{1}$ is asymptotically most correlated with $A_{2}$ if and only if Equation~\ref{mainCondition} is satisfied for $q=1$.

We expand the correlation between $A_{1}$ and $A_{3}$ as follows:
\begin{align*}
\rho(A_{1}, A_{3}) &= \frac{\sum_{i=1}^{n}(A_{i1}A_{i3})}{\sqrt{\sum_{i=1}^{n}A_{i1}^{2} \sum_{i=1}^{n}A_{i3}^{2}}} \\
 &= \frac{\sum_{i=1}^{n}(A_{i1}A_{i3}/n)}{\sqrt{\sum_{i=1}^{n}(A_{i1}/n) \sum_{i=1}^{n}(A_{i3}/n)}} \\
 & \stackrel{a.s.}{\rightarrow} \frac{\sum_{k=1}^{K} (\pi_{k} B_{k1}B_{k2})}{\sqrt{ \sum_{k=1}^{K} (\pi_{k} B_{k1}) \sum_{k=1}^{K} (\pi_{k} B_{k2})}} \\
 &= \frac{E(Q_{1}Q_{2})}{\sqrt{E(Q_{1})E(Q_{2})}},
\end{align*}
where the first line is by the definition of correlation, the second line follows by noting that the entries of $A$ are $0$ and $1$, the third line follows by passing to limit, the fourth line simplifies the expression by our definition of $\{Q_{q}\}$. Note that we assumed known class-membership for the first three vertices, but they do not affect the asymptotic correlation and thus not considered in the limit expression.

By a similar expansion, we have 
\begin{align*}
\rho(A_{1}, A_{2}) \stackrel{a.s.}{\rightarrow} \frac{E(Q_{1}^{2})}{E(Q_{1})}. 
\end{align*}

So $A_{1}$ is asymptotically most correlated with $A_{2}$ if and only if 
\begin{align*}
& \frac{E(Q_{1}^{2})}{E(Q_{1})} > \frac{E(Q_{1}Q_{2})}{\sqrt{E(Q_{1})E(Q_{2})}} \\
\Leftrightarrow & \frac{E(Q_{1}^{2})}{E(Q_{1})} > \frac{\rho_{12} \sqrt{E(Q_{1}^{2})E(Q_{2}^{2})}}{\sqrt{E(Q_{1})E(Q_{2})}} \\
\Leftrightarrow & \sqrt{\frac{E(Q_{2})}{E(Q_{1})}} > \rho_{12} \sqrt{\frac{E(Q_{2}^{2})}{E(Q_{1}^{2})}} \\ 
\Leftrightarrow & \rho_{12}^{2} \cdot \frac{E(Q_{2}^{2})}{E(Q_{1}^{2})} < \frac{E(Q_{2})}{E(Q_{1})}.
\end{align*}
The above derivation is always valid when class $2$ is replaced by any class $r \neq 1$. Thus we proved Lemma~\ref{SBMLemma1}.  
\end{proof}

\subsubsection{Proof of Lemma~\ref{SBMLemma2}}




\begin{proof}
Without loss of generality, denote $\alpha$ and $A_{s+1}$ as two adjacency columns from class $1$, $A_{(s)}=[A_{1}|\cdots|A_{s}]$, and $C=[c_{1}, \cdots, c_{s}]$. It suffices to prove that as $n \rightarrow \infty$, we always have
\begin{equation}
\label{equation1}
\rho(\alpha, A_{s+1}) > \rho(\alpha, C \cdot A_{(s)}) = \rho(\alpha, \sum_{j=1}^{s}c_{j}A_{j})
\end{equation}
for any non-negative vector $C$, and all possible $A_{(s)}$ whose columns are not of class $1$. Note that Lemma~\ref{SBMLemma1} show that Equation~\ref{mainCondition} is sufficient and necessary for Equation~\ref{equation1} to hold at $s=1$ for any $C$; and the same condition is still sufficient for Equation~\ref{equation1} to hold at any $s \geq 1$, under the additional assumption that $C$ is a non-negative vector; when $s=0$, Equation~\ref{equation1} trivially holds.

In the proof of Lemma~\ref{SBMLemma1}, we already showed that 
\begin{align*}
\rho(\alpha, A_{s+1}) \stackrel{a.s.}{\rightarrow} \frac{E(Q_{1}^{2})}{E(Q_{1})}.
\end{align*}
Next let us expand $\rho(\alpha, \sum_{j=1}^{s}c_{j}A_{j})$ as follows:
\begin{align*}
\rho(\alpha, C \cdot A_{(s)}) & = \rho(\alpha, \sum_{j=1}^{s}c_{j}A_{j}) \\
& = \frac{\sum_{i=1}^{n}(\alpha_{i}(\sum_{j=1}^{s} c_{j} A_{ij}))}{\sqrt{\sum_{i=1}^{n}\alpha_{i}^{2} \sum_{i=1}^{n}(\sum_{j=1}^{s} c_{j} A_{ij})^{2}}} \\
& = \frac{\sum_{j=1}^{s} c_{j} (\sum_{i=1}^{n} \alpha_{i} A_{ij}/n)}{\sqrt{\sum_{i=1}^{n}(\alpha_{i}^{2}/n) \sum_{i=1}^{n}(\sum_{j=1}^{s} c_{j} A_{ij})^{2}/n}} \\
& \leq \frac{\sum_{j=1}^{s} c_{j} (\sum_{i=1}^{n} \alpha_{i} A_{ij}/n)}{\sqrt{\sum_{i=1}^{n}(\alpha_{i}/n) \sum_{i=1}^{n}(\sum_{j=1}^{s} c_{j}^{2} A_{ij}/n)}} \\
& \stackrel{a.s.}{\rightarrow} \frac{\sum_{j=1}^{s}(c_{j} \sum_{k=1}^{K} (\pi_{k} B_{k1}B_{ky_{j}}))}{\sqrt{ \sum_{k=1}^{K} (\pi_{k} B_{k1}) \sum_{j=1}^{s}(c_{j}^{2} \sum_{k=1}^{K} (\pi_{k} B_{ky_{j}}))}} \\
& = \frac{\sum_{j=1}^{s} c_{j}E(Q_{1}Q_{y_{j}})}{\sqrt{\sum_{j=1}^{s}c_{j}^2 E(Q_{1})E(Q_{y_{j}})}},
\end{align*}
where $y_{j}$ denotes the class membership for $A_{j}$. All other steps being routine, the inequality in the above expansion is due to $\sum_{j=1}^{s} (c_{j} A_{ij})^{2} \geq \sum_{j=1}^{s} c_{j}^{2} A_{ij}$, which is obvious because $c_{j}$ and $A_{ij}$ are always non-negative.

Therefore, in order to show Equation~\ref{equation1} holds asymptotically, it suffices to prove that 
\begin{align*}
& \frac{E(Q_{1}^{2})}{E(Q_{1})} > \frac{\sum_{j=1}^{s} c_{j}E(Q_{1}Q_{y_{j}})}{\sqrt{\sum_{j=1}^{s}c_{j}^2 E(Q_{1})E(Q_{y_{j}})}} \\
\Leftrightarrow & \frac{E(Q_{1}^{2})}{E(Q_{1})} > \frac{\sum_{j=1}^{s} \rho_{1y_{j}}c_{j} \sqrt{E(Q_{1}^{2})E(Q_{y_{j}}^{2})}}{\sqrt{\sum_{j=1}^{s}c_{j}^2 E(Q_{1})E(Q_{y_{j}})}} \\
\Leftrightarrow & \sqrt{\sum_{j=1}^{s}c_{j}^2 \frac{E(Q_{y_{j}})}{E(Q_{1})}} > \sum_{j=1}^{s} \rho_{1y_{j}}c_{j} \sqrt{\frac{E(Q_{y_{j}}^{2})}{E(Q_{1}^{2})}}.
\end{align*}
The last inequality holds when 
\begin{align*}
\rho_{1y_{j}}^{2} \cdot \frac{E(Q_{y_{j}}^{2})}{E(Q_{1}^{2})} < \frac{E(Q_{y_{j}})}{E(Q_{1})},
\end{align*}
which is exactly Equation~\ref{mainCondition} when $y_{j} \neq 1$. 

Therefore, Equation~\ref{mainCondition} and non-negative $C$ are sufficient for Lemma~\ref{SBMLemma2} to hold.
\end{proof}

\subsubsection{Proof of Theorem~\ref{SBMTheorem}}



\begin{proof}
In order to prove that SRC is a consistent classifier with $L_{n} \rightarrow 0$, it suffices to prove that the adjacency matrix generated by SBM satisfies a principal angle condition in \cite{ShenChenPriebe2015}. This consistency holds for either $\ell 1$ minimization or orthogonal matching pursuit at any $s \geq 1$, assuming the sparse coefficient $x$ is non-negative. 

For the adjacency matrix under SBM, suppose $\alpha$ is a fixed adjacency column of class $q$, $A_{s+1}$ is a random adjacency column of class $q$, and $A_{(s)}$ is a random matrix whose columns are not of class $q$. Then to show $L_{n} \rightarrow 0$ at any $s$, the principal angle condition requires that $\theta(\alpha, A_{s+1}) < \theta(\alpha, A_{(s)})$, for all possible $\alpha$ under SBM.

The principal angle condition is used in two areas of SRC: first it guarantees that the selected sub-matrix by SRC contains at least one observation of the correct class; second it guarantees the sparse coefficient with respect to the correct class dominates the sparse representation. Then if data is always non-negative (which always holds for graph adjacency) and the sparse representation is non-negative (which can be relaxed to bounded below in \cite{ShenChenPriebe2015}), such dominance is sufficient for the correct classification of SRC.

Lemma~\ref{SBMLemma1} proves the principal angle condition at $s=1$, which is sufficient for the first point above. Lemma~\ref{SBMLemma2} proves the principal angle condition at any $s$, which is equivalent to the second point above under the non-negative constraint. Therefore we establish SRC consistency for SBM.

Note that there are two small differences: First, the original condition is not in the limit form, while Lemma~\ref{SBMLemma1} and Lemma~\ref{SBMLemma2} are proved asymptotically for SBM; this change has no effect for classification consistency. Second, in \cite{ShenChenPriebe2015} we separate the principal angle condition from the non-negative constraint, while in Lemma~\ref{SBMLemma2} we effectively combine the non-negative constraint into proving the principal angle condition; this does not affect the result either, because when the sparse coefficient are constrained to be non-negative, the principal angle between two subspaces are also constrained accordingly.
\end{proof}

\section*{Acknowledgements}
This work is partially supported by a National Security Science and Engineering Faculty Fellowship (NSSEFF), Johns Hopkins University Human Language Technology Center of Excellence (JHU HLT COE), and the XDATA
program of the Defense Advanced Research Projects Agency (DARPA) administered through Air Force Research Laboratory contract FA8750-12-2-0303. We would also like to thank Minh Tang for his thoughtful discussions.

\bibliographystyle{IEEEtran}
\bibliography{robust_est}

\end{document}